\documentclass{elsarticle}  
\usepackage{graphicx}
\usepackage{latexsym}
\usepackage{amsmath}
\usepackage{amssymb}
\usepackage{hyperref}
\usepackage{geometry}
\usepackage{makecell}

\usepackage{framed,multirow}
\usepackage{xcolor}
\usepackage{colortbl}
\usepackage{subcaption}
\usepackage[ruled, lined, commentsnumbered, longend]{algorithm2e}

\def\picScaleFr{0.23}
\def\picScaleF{0.18}
\def\picScaleDP{0.43}
\def\picScaleA{0.29}
\def\picScaleB{0.155}
\def\picScaleCa{0.053}
\def\picScaleCb{0.056}
\def\picScaleCc{0.052}
\def\picScaleEc{0.23}
\DeclareMathOperator*{\argmin}{arg\,min}

\newcolumntype{u}{>{\columncolor{white}}l}
\newcolumntype{v}{>{\columncolor{white}}c}
\newcolumntype{w}{>{\columncolor{white}}r}

\begin{document}

\begin{frontmatter}

\title{OsmLocator: locating overlapping scatter marks with a non-training generative perspective}

\author[cigit,ugent]{Yuming Qiu \corref{cor1}}
\ead{qiuym@cigit.ac.cn or ccqiuym@126.com}
\author[ugent]{Aleksandra Pizurica}
\ead{aleksandra.pizurica@ugent.be}
\author[ugent]{Qi Ming}
\ead{chaser.ming@gmail.com}
\author[ugent]{Nicolas Nadisic}
\ead{nicolas.nadisic@ugent.be}

\affiliation[cigit]{organization={Chongqing Institute of Green and Intelligent Technology, Chinese Academy of Sciences},
		addressline={No.266, Fangzheng Ave.},
		city={Chongqing},
		postcode={400714},
		country={China}}
\affiliation[ugent]{organization={GAIM, Faculty of Engineering and Architecture, Ghent University},
		addressline={Sint-Pietersnieuwstraat 25},
		city={Gent},
		postcode={9000},
		country={Belgium}}
\cortext[cor1]{Corresponding author}
		
\begin{abstract}
Automated mark localization in scatter images, greatly helpful for discovering knowledge and understanding enormous document images and reasoning in visual question answering AI systems, is a highly challenging problem because of the ubiquity of overlapping marks. Locating overlapping marks faces many difficulties such as no texture, less contextual information, hallow shape and tiny size. Here, we formulate it as a combinatorial optimization problem on clustering-based re-visualization from a non-training generative perspective, to locate scatter marks by finding the status of multi-variables when an objective function reaches a minimum. The objective function is constructed on difference between binarized scatter images and corresponding generated re-visualization based on their clustering. Fundamentally, re-visualization tries to generate a new scatter graph only taking a rasterized scatter image as an input, and clustering is employed to provide the information for such re-visualization. This method could stably locate severely-overlapping, variable-size and variable-shape marks in scatter images without dependence of any training dataset or reference. Meanwhile, we propose an adaptive variant of simulated annealing which can works on various connected regions. In addition, we especially built a dataset named SML2023 containing hundreds of scatter images with different markers and various levels of overlapping severity, and tested the proposed method and compared it to existing methods. The results show that it can accurately locate most marks in scatter images with different overlapping severity and marker types, with about 0.3 absolute increase on an assignment-cost-based metric in comparison with state-of-the-art methods. This work is of value to data mining on massive web pages and literatures, and shedding new light on image measurement such as bubble counting. Code sources and SML2023 dataset are available in  \href{https://github.com/ccqym/OsmLocator}{https://github.com/ccqym/OsmLocator}.
\end{abstract}
\end{frontmatter}

\section{Introduction} \label{sec_intr}
Understanding figures in enormously-existing and fast-growing scientific publications and web pages is becoming a significant challenge for artificial intelligence. For examples, discovering knowledge or mining data from various charts \cite{davila_chart_2021,chart_info_2023}, or reasoning in visual question answering (VQA) AI systems \cite{FigureQA2017,DVQA2018cvpr,PlotQA2020wacv} especially working with generative pre-trained models such as ChatGPT. 
Scattering is a vital visualization method for data presentation \cite{Cleveland_1985,friendly_early_2005}. Various rasterized scatter images such as scatter plots, scatter graphs, scatter maps or scatter diagrams, commonly exist in these valuable professional documents. It was estimated that between 70 and 80 percent of graphs used in scientific publications are scatter plots \cite{friendly_early_2005}. Unfortunately, publications usually do not expose the underlying data, and raw data in scatter images have been lost gradually with time elapsing. VQA systems need to answer questions only with scatter images. So, extracting valuable data in scatter images becomes increasingly significant and indispensable. In such data extraction task, the first and one of the most challenging parts is to locate all marks in scatter images.

\begin{figure}[!ht]
	\centering
	\setlength{\tabcolsep}{0.03\columnwidth}
	\begin{tabular}{uw}
		\includegraphics[width=\picScaleDP\columnwidth]{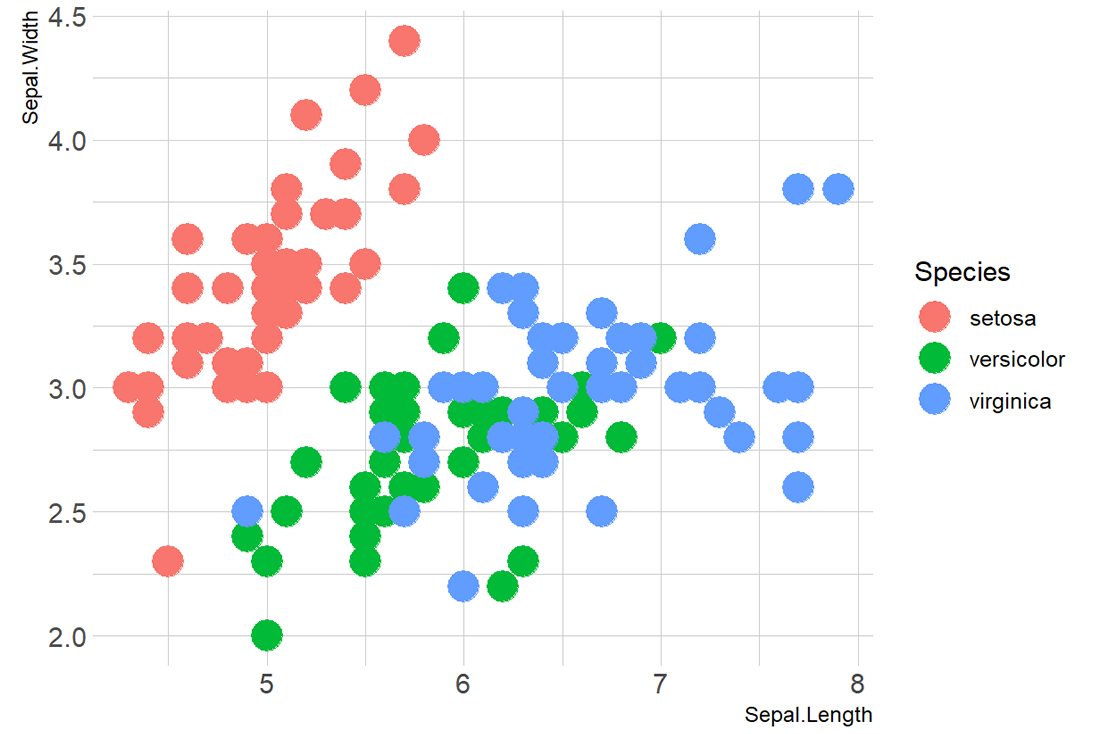}&
		\includegraphics[width=\picScaleDP\columnwidth]{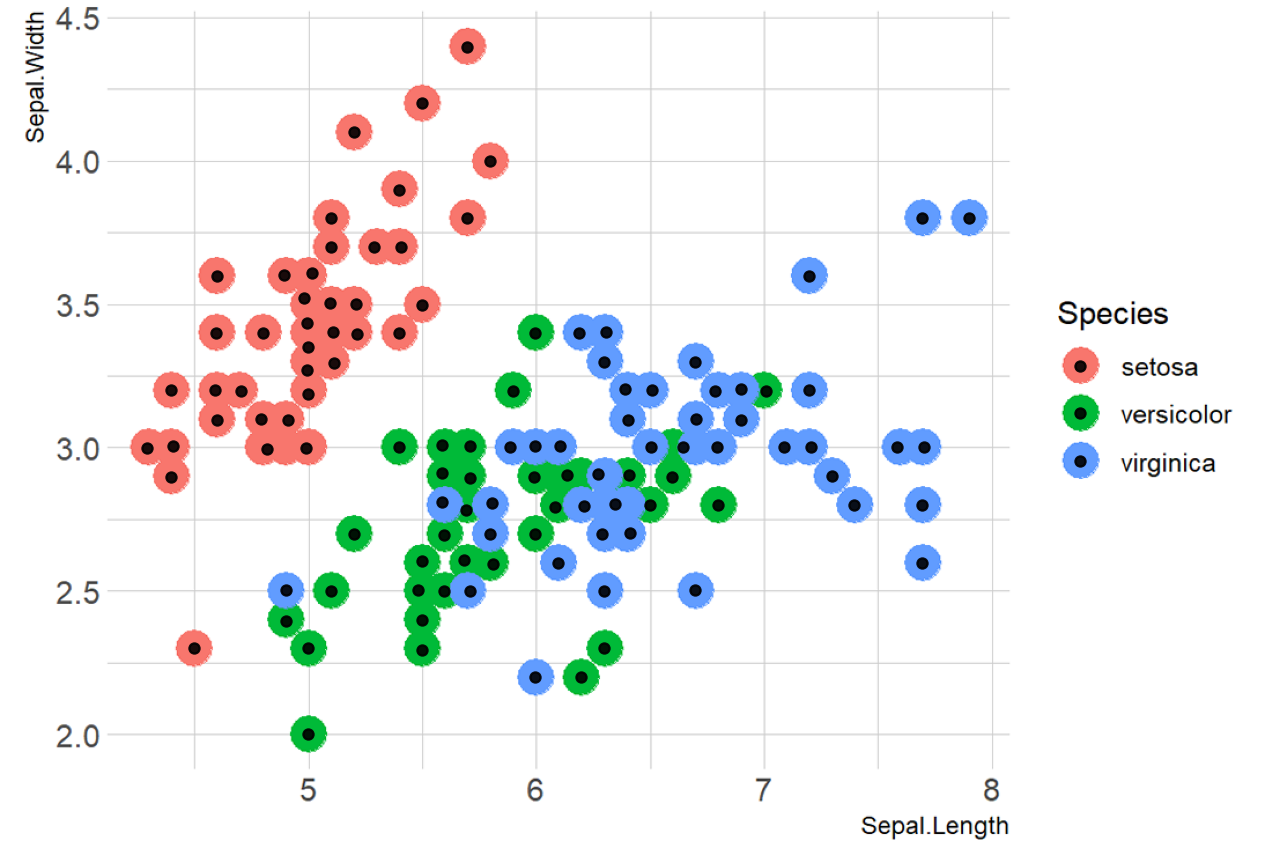}\\
		\includegraphics[width=\picScaleDP\columnwidth]{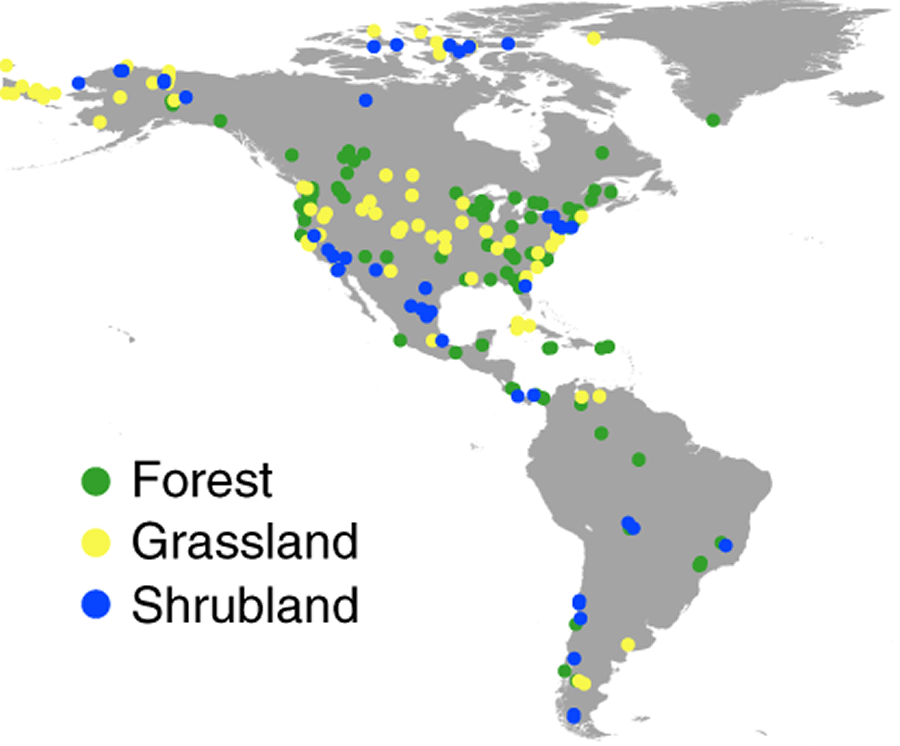}&
		\includegraphics[width=\picScaleDP\columnwidth]{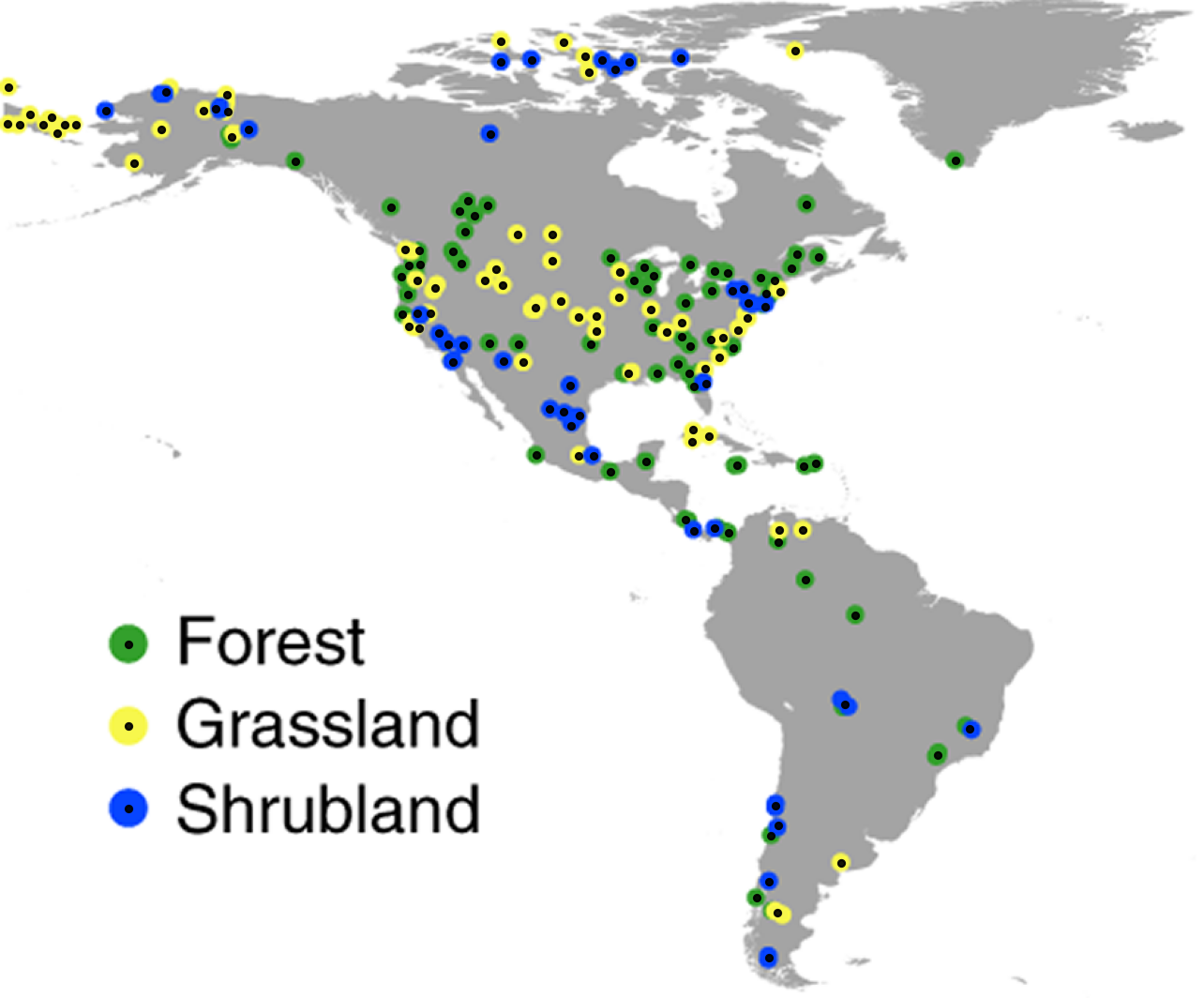}
	\end{tabular}
	\caption{Examples of scatter mark localization from a scatter plot and a scatter map. The left column lists the input scatter images. The right column shows the located marks (black points) by our method.\label{fig:intro_show}}
\end{figure}

It is extremely common that many marks are overlapped together intentionally to form connected image regions to demonstrate the patterns in data points and reflect the data dispersion degree. But this brings about grand challenges to locate those overlapping marks for uncovering underlying data points. First, the connected regions formed by amount of marks usually do not have texture which is very important for many machine-learning methods. Second, overlapping marks in a connected region have less contextual information to leverage. Third, scatter marks usually are very tiny, and this means that many methods depends on convolution or pool operation may face disappeared reception field. Forth, there are many types of marker such as square, circle, diamond and star to draw the scatter images and many marks are hallow, only formed marks with different edge colour. The images in the left column of Figure \ref{fig:intro_show} show the difficulty to locate all marks even for human vision.
The existing methods for mark localization can be categorized into three classes: 1) manual, such as the R package \textit{digitize} \cite{poisot_digitize_2011}; 2) semi-automated, such as ScatterPlotAnalyzer \cite{paszynski_scatterplotanalyzer_2021} and \cite{Sreevalsan2021}; 3) automated, such as ScatterScanner \cite{baucom_scatterscanner_2013}, Scatteract \cite{cliche_scatteract_2017} and Chartem \cite{Jiayun2021}. These works made efforts on mark localization or extraction, but almost all of them left overlapping marks localization as unsolved.

In this paper, we propose a new non-training generative way to locate overlapping marks in scatter images without any training dataset or reference and developed a tool named \textit{OsmLocator} for community. Overlapping marks localization is formulated as an optimization problem on clustering-based re-visualization, and the goal is converted to find the status of multi-variables when an objective function reaches a minimum. The objective function of optimization is mainly constructed on the difference between binarized scatter images and corresponding generated re-visualization based on the clustering results of such binarized scatter images. The underlying idea is to find the best way to generate the scatter graph to make it agree with the original visualization as perfect as possible for data in a scatter image. The way we found is to leverage clustering algorithms to provide the necessary information for marks re-visualization, and employ simulated annealing to find the optimal solution for how to cluster and re-visualize. The objective function we constructed has stable performance on different input scales with same control parameters. Moreover, we propose an adaptive variant of simulated annealing which can works on various connected regions.

In order to validate the performance of the proposed method, we specially built a new dataset called SML2023 for the scatter mark localization task. This dataset contains 297 synthetic scatter images with different markers and various levels of overlapping severity. We tested the proposed method and compared it with existing methods. The experimental results show that it can locate most marks in scatter images for different overlapping severity and marker types. Our \textit{OsmLocator} achieves state-of-the-art performance, with about 0.3 absolute increase on an assignment-cost-based metric in comparison with state-of-the-art methods. The right column in Figure \ref{fig:intro_show} presents two examples of the performance of our method for locating marks in a scatter plot from \cite{Holtz_2021} and a scatter map from \cite{Haozhi2021}. More over, we made experiments to discuss the influence of control parameters and the effectiveness and efficiency of simulated annealing. This work is of value to data mining on massive web pages and literatures, and shedding new light on image measurement such as bubble counting.
The contributions of this work are four-fold:
\begin{itemize}
	\item We take a new non-training generative perspective to formulate the overlapping scatter marks localization task as an optimization problem to estimate the number of clusters and the marker type to find the best re-visualization agreeing with original visualization.
	\item A new self-adaptive variant of Simulated Annealing is proposed to find the optimal solution to locate overlapping marks.
	\item A new synthetic dataset was built to validate various methods for overlapping marks localization in community.
	\item We implemented the proposed method in python and tested it on the new dataset; the experimental results show that it can locate most marks in scatter images with different overlapping severity and marker types,
	with about 30\% absolute increase on an assignment-cost-based metric in comparison with state-of-the-art methods.
\end{itemize}

The rest of this paper is organized as follows. Section \ref{sec_prior} firstly reviews related works. Then the details of proposed method are presented in Section \ref{sec_method}. Next, we introduce datasets and metric in Section \ref{subsec_dataset}, and report experimental results in Section \ref{sec_expr}. At last, conclusions are drawn in Section \ref{sec_conc} after some discussion in Section \ref{sec_discz}.

\section{Related works} \label{sec_prior}
Analysing figures or charts in publications and web pages is a valuable but challenging task. Many works \cite{Sagnik_2013,Praczyk_2013,Ray_2015,Choudhury_2015,Siegel_2016,Choudhury_2016,Clark_2016,Noah_2018,Chai2021,Jiayun2021,Alexiou_2021,Brandt_2021,Zhou2021,Filip2022,Derya2022,Kato2022,Mishra2022,Mishra_2022} have devoted to automated chart analysis. Some works \cite{Yan_2013,davila_chart_2021,shahira_towards_2021} also have made comprehensive survey. Scatter plot analysis is a fundamental step for chart mining \cite{davila_chart_2021}, infographics \cite{chart_info_2023}, visual reasoning \cite{FigureQA2017} or plot reasoning \cite{PlotQA2020wacv} or visualization understanding \cite{DVQA2018cvpr}. Except manual methods like \cite{poisot_digitize_2011,crowdchart_Chai_2021}, there are roughly five ways for semi-automated or automated mark localization.

\paragraph{Filter-based methods}
usually utilize a pre-prepared kernel to filter scatter images and then locate marks by finding some kind of pattern in filtering results. ScatterScanner \cite{baucom_scatterscanner_2013} employs a fixed Gaussian kernel whose size from isolated single marks to filter a scatter image, and then locate marks by finding peaks in the filtered map. It can locate a part of connected marks in special positions, but usually fails to recall many connected or overlapping marks. Furthermore, it is very sensitive to the kernel size, usually missing the marks whose size is different from the kernel's. Some figures in section \ref{sec_expr} presents some experimental results by it.

\paragraph{Supervised machine learning based methods} \label{subsec_ML}
use machine learning models and annotated datasets to predict the mark positions and types. A recent effort is Scatteract \cite{cliche_scatteract_2017}. It formulates mark detection as an object detection problem, and employed a deep learning model ReInspect \cite{stewart_end_end_2016} to detect scatter marks. Although it can recall some overlapping marks in different colours, it still failed on many overlapping marks. So far, handling overlapping objects is the most difficult case in object detection \cite{Tausif2023}, because there are some difficulties limiting the effectiveness of deep learning methods, such as no texture, less contextual information, hallow shape and tiny size.

\paragraph{Placement based methods}
treat mark localization as a problem about location estimation to place specified shapes into correct positions. For each point-type shape, the work \cite{browuer_segregating_2008} uses simulated annealing to find the optimal position and with an additional swapping marker type. They took almost all positions in an input image and marker types as the search space, resulting in a huge search space and may take a huge number of iterations. It also depends on and is sensitive to the input binarized shape, limiting its application.
Notably, we also employ simulated annealing in our method, but they are completely different technical routes because the problem formulation, objective functions, dependent inputs and the scale of search space are different.

\paragraph{Curvature based methods}
suppose that the curvature of break points (connecting points) in the object contour always reach a local maximum value. This kind of methods usually applied to measure bubble size distribution or count number of bubbles, such as \cite{ZhangWH2012,Lau2013,ShengZhong2016,ZouTong2021}. However, it is inappropriate to leverage them into scatter mark localization. First, they are mainly used for elliptical objects, difficult for many types of overlapping marks such as $\times$ and $\vartriangle$. Second, they usually depend on the quality of concave detection; there are many cases without break points in the contour of overlapping marks in scatter images as shown in Figures \ref{fig:intro_show} and \ref{fig:dataset_examples}. Third, the problem to find an optimal ellipse subset usually resort to solve an integer programming problem, whose computation required increases exponentially with the size of contour segments.

\paragraph{Watermarking based methods}
pre-embed information in data-visualization phase, and then use such embedded information to help to recover data. \cite{Mark_2005} proposed a watermarking scheme for map and chart images by modifying boundaries of homogeneous color components. Chartem \cite{Jiayun2021} embeds information into the background region while keeping foreground components unchanged, and then uses an extractor to extract embedded information from chart images. This method require embedding enough information when making charts and may affect readers' experience and suffer from limited background area, distortions and resizing.

\paragraph{Tensor-fields-based methods}
usually use the locality of the boundary of scatter marks in tensor fields to analyse scatter marks \cite{paszynski_scatterplotanalyzer_2021,Sreevalsan2021}, but they depends on prior knowledge (the parameters of DBSCAN clustering) specified for every image to separate scatter marks. So they are semi-automated. 

All in all, many studies had been devoted into the mark localization problem, but segregating or locating overlapping marks remains unsolved.

\section{Method} \label{sec_method}	
\subsection{Overview}
\begin{figure*}[!ht]
	\centering
	\includegraphics[width=1.0\textwidth]{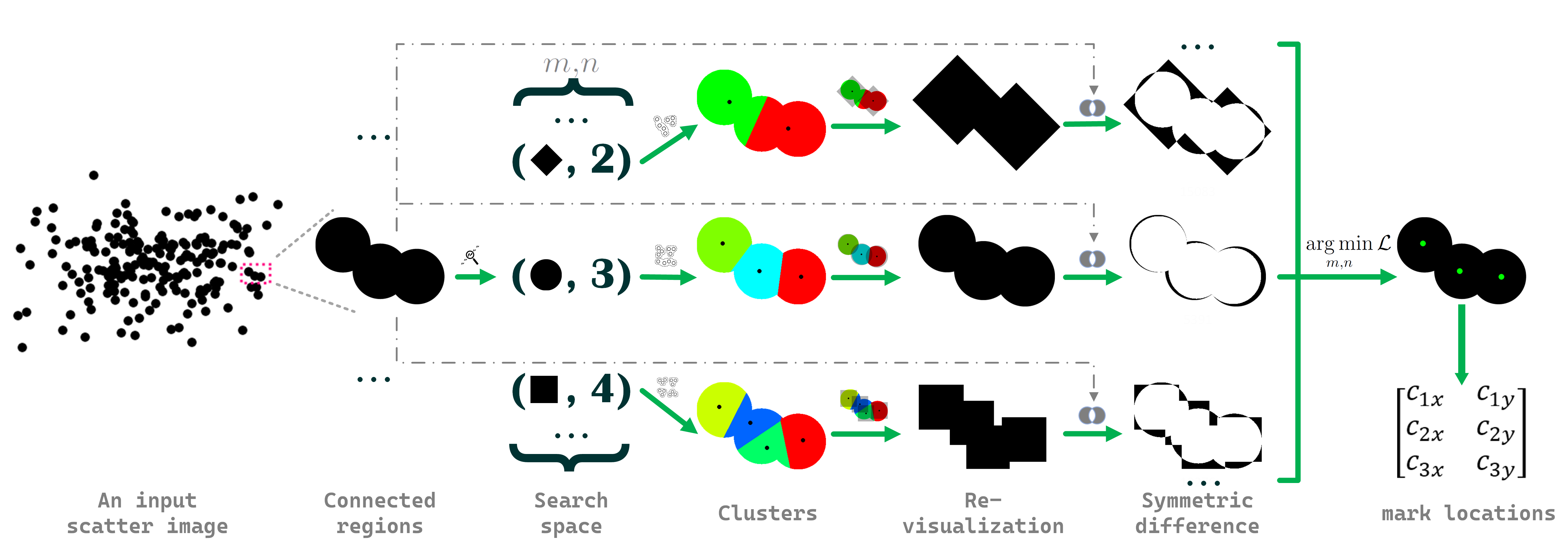}
	\caption{An illustration of the proposed methodology. For each connected region in a binarized scatter image, firstly a solution consisting of number $n$ of clusters and a supportable marker type $m$ is generated; then all pixels are clustered into $n$ groups and a re-visualization is made based on $n$ and $m$; next, the value of a loss function $\mathcal{L}$ is calculated based on the symmetric difference between the re-visualization and the connected region; at last, the mark locations and marker type are obtained by resolving $\argmin\mathcal{L}$ through optimization.} \label{fig_underlying_idea}
\end{figure*}
In this work, we take a new non-training generative way to formulate the mark localization task as a combinatorial optimization problem to locate scatter marks by minimizing an objective function which is mainly defined on the difference between a connected region and a re-visualization generated on the clustering results of such connected region. The underlying idea is illustrated in Figure \ref{fig_underlying_idea}. A basic hypothesis is that a perfect generated re-visualization should be exactly same as the original visualization if we can find the exact geometric centroid and the right marker type used when the original input scatter image was drawn. In turn, if we can find the perfect re-visualization, it is feasible to get the data points represented by geometric centroid of marks. The real difficulty is how to find the perfect re-visualization using only scatter images consisting of numerous overlapping marks. Our solution for this is to leverage clustering to group a connected region into a certain amount of clusters by the Euclidean distance if treating each pixel of the connected region as a "data point" with two features namely its coordinates in the image. Figure \ref{fig_underlying_idea} vividly shows that the re-visualization can best coincide with the original scatter images when the marker type and the number of clusters are correct. However, the subsequent questions become intractable: how to determine the number of clusters? what properties of marks used in the original drawing phase? and do re-visualization and original visualization have one-to-one correspondence? 
For them, we proposed to use combinatorial optimization to formulate and resolve these questions.

The overview procedure of this method is as follows: firstly, an input scatter image is broken down into multiple connected regions in order to reduce the problem scale, in which some of them denote single marks but others are formed by overlapping marks. Then, for each connected region, a search space including finite types of marks and quantities of clusters is settled according to size of the connected regions. Next, for a tuple of a marker type and a number of clusters, a re-visualization is generated based on the clustering results and the marker type; and subsequently the symmetric difference between the re-visualization and corresponding connected region is calculated. At last, the marker type and mark locations are obtained by resolving the $\argmin$ of an objective function defined on the symmetric difference and other constraints. The following subsections will expand the details.

\subsection{Re-visualization on clusters} \label{sec:revis_on_clst}
Drawing a scatter image is mainly controlled by two parts: 1) coordinate information showing the location of data and relationship among them; 2) a marker with type, size, colour properties controlling the appearance of marks. Likewise, a re-visualization also need these information. So, the key to mark localization is to find coordinate information and marker style only based on a rasterized scatter image.

In this work, we employ clustering algorithms to obtain coordinates and marker size, and then look for the possibly best cluster division and marker type via optimization. Formally, given a number of clusters $n$ and a marker type $m$ for a connected region in a binary scatter image, the position of all pixels firstly are collected into a set; then a clustering algorithm, for example K-means, is used to cluster the set into $n$ groups; Next, for each cluster, a new mark is drawn in the centroid of the cluster by the marker type $m$ and the radius $r$ calculated on the cluster. A principle of re-visualization is to coincide with the marks in original scatter image as closely as possible. The radius of a cluster $r$ is calculated as 
\begin{equation}
	r = \max_{p\epsilon \mathbb{A}}{||\vec{p}-\vec{c}||}, \label{eqRadius}
\end{equation}
where $p$ denotes the coordinate vector of each pixel in a cluster $\mathbb{A}$, and $c$ denotes the centroid's coordinate of the cluster $\mathbb{A}$.

\begin{figure}[!h]
	\centering
	\small
	\setlength{\tabcolsep}{0.008\columnwidth}
	\begin{tabular}{cccccc}
		\includegraphics[width=\picScaleB\columnwidth]{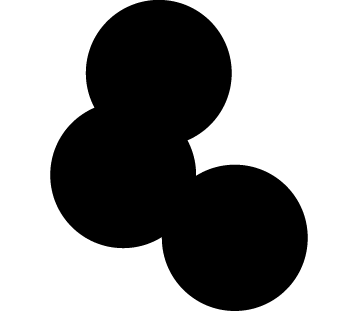}&
		\includegraphics[width=\picScaleB\columnwidth]{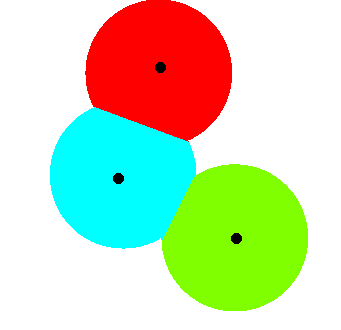}&
		\includegraphics[width=\picScaleB\columnwidth]{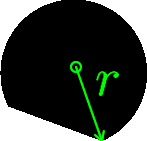}&
		\includegraphics[width=\picScaleB\columnwidth]{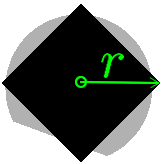}&
		\includegraphics[width=\picScaleB\columnwidth]{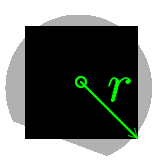}&
		\includegraphics[width=\picScaleB\columnwidth]{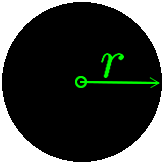}\\
		Input & 3-clusters & A cluster & Diamond & Square & circle
	\end{tabular}
	\caption{Demonstration of re-visualizations with three different markers and a same radius $r$ of a cluster. \label{fig_revis_radius}}
\end{figure}
For different marker types, cluster radii may play different roles in re-visualization. Let us take three commonly-used markers as examples. For type "circle", a cluster radius is the radius of a circle; For types "square" and "diamond", the cluster radius is a half of the diagonal. 
Figure \ref{fig_revis_radius} shows an example of re-visualizations for a cluster using three typical makers with a same radius.

\begin{figure}[h]
	\centering
	\setlength{\tabcolsep}{5pt}
	\begin{tabular}{u|vvvv|vvvv|vvvw}
		\includegraphics[scale=0.092]{imgs/method/circle_ex.png}&
		\includegraphics[scale=\picScaleCa]{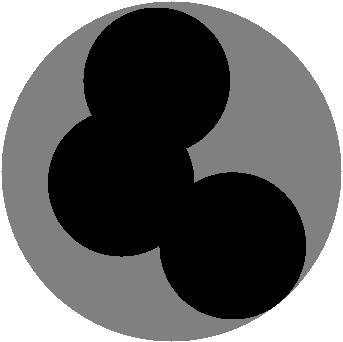}&
		\includegraphics[scale=\picScaleCa]{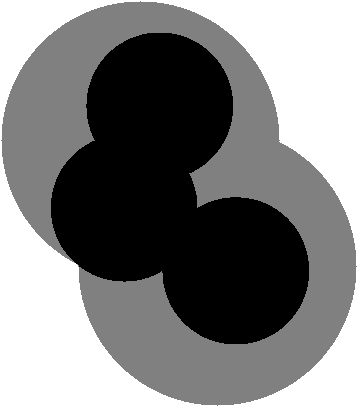}&
		\includegraphics[scale=\picScaleCa]{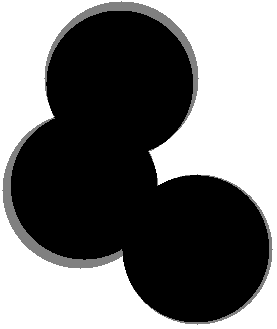}&
		\includegraphics[scale=\picScaleCa]{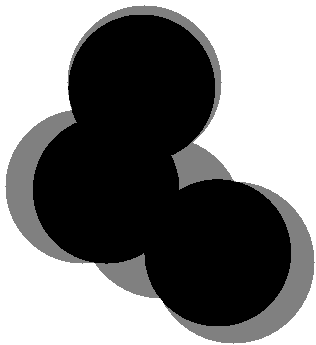}&
		\includegraphics[scale=\picScaleCa]{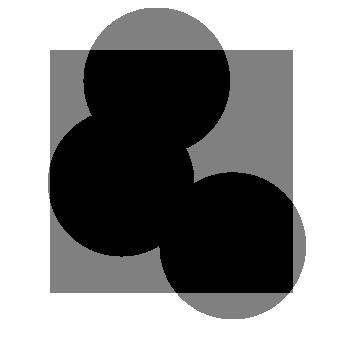}&
		\includegraphics[scale=\picScaleCa]{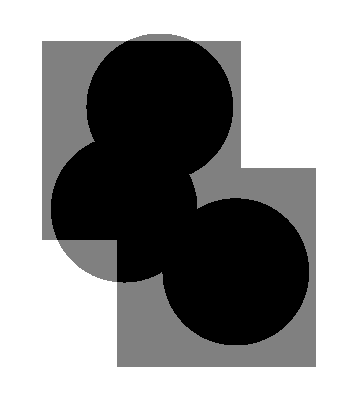}&
		\includegraphics[scale=\picScaleCa]{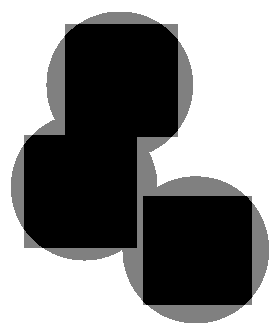}&
		\includegraphics[scale=\picScaleCa]{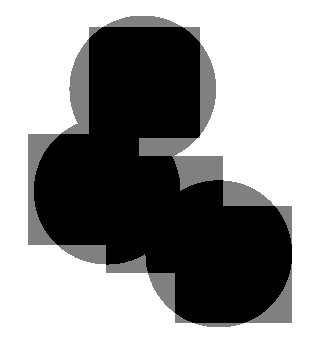}&
		\includegraphics[scale=\picScaleCa]{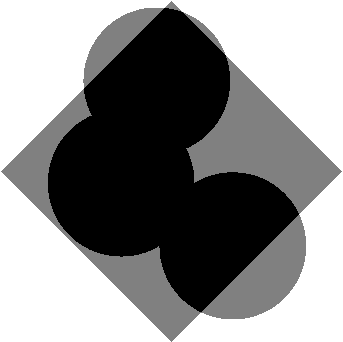}&
		\includegraphics[scale=\picScaleCa]{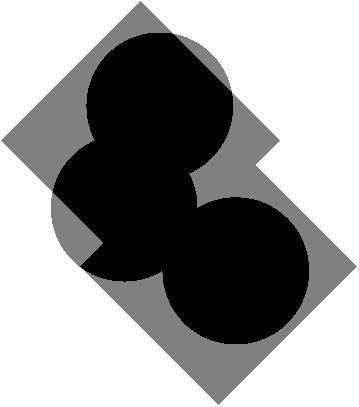}&
		\includegraphics[scale=\picScaleCa]{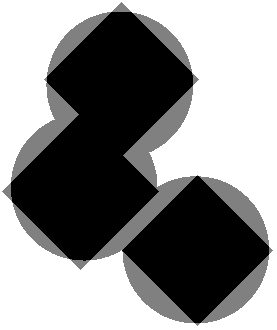}&
		\includegraphics[scale=\picScaleCa]{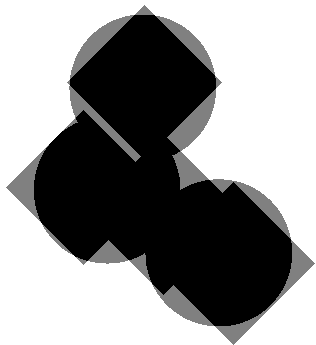}\\
		\includegraphics[scale=0.072]{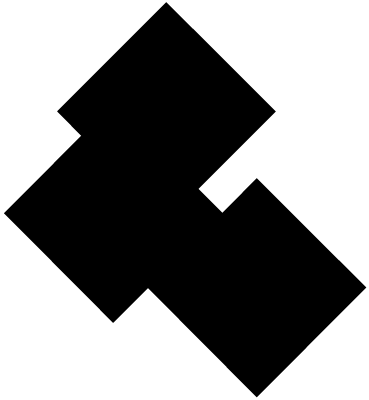}&
		\includegraphics[scale=\picScaleCb]{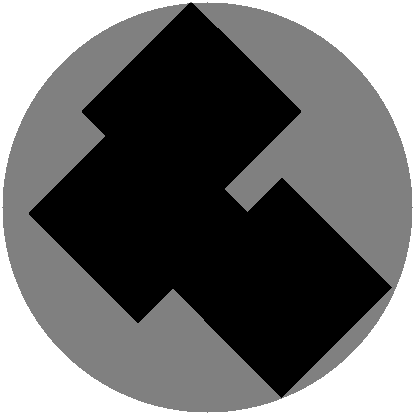}&
		\includegraphics[scale=\picScaleCb]{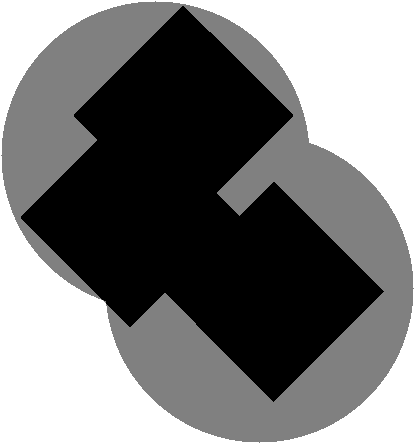}&
		\includegraphics[scale=\picScaleCb]{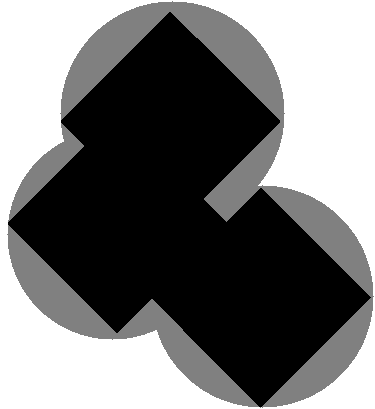}&
		\includegraphics[scale=\picScaleCb]{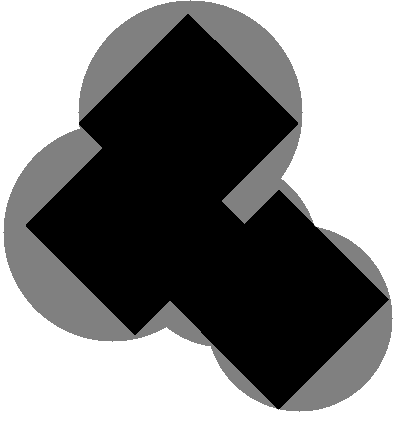}&
		\includegraphics[scale=\picScaleCb]{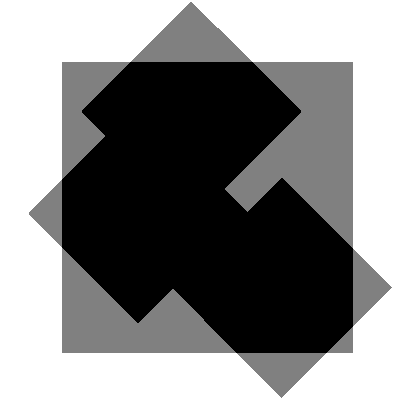}&
		\includegraphics[scale=\picScaleCb]{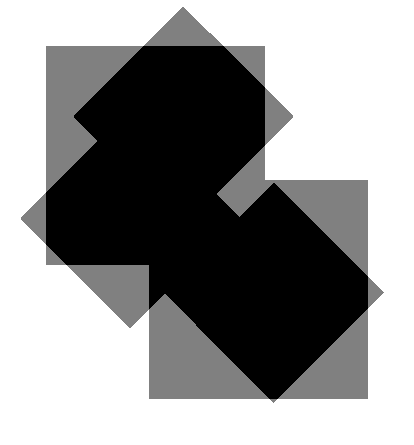}&
		\includegraphics[scale=\picScaleCb]{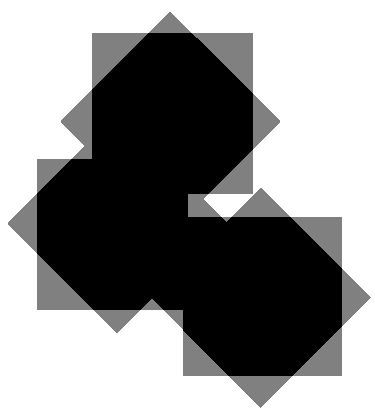}&
		\includegraphics[scale=\picScaleCb]{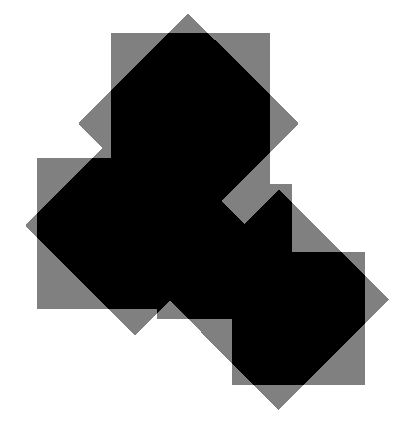}&
		\includegraphics[scale=\picScaleCb]{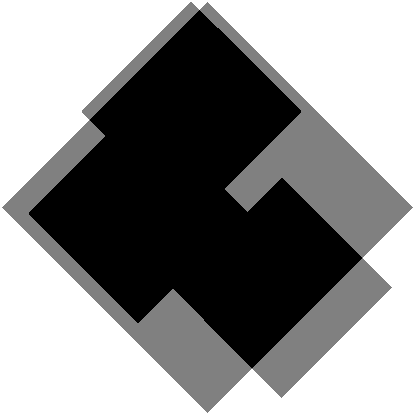}&
		\includegraphics[scale=\picScaleCb]{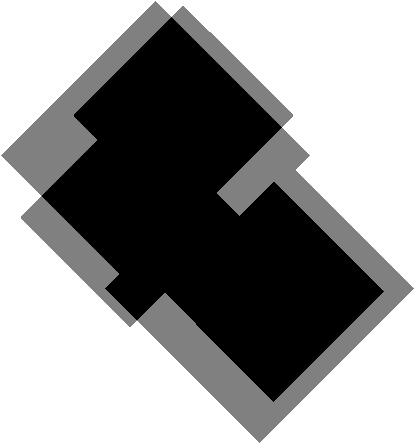}&
		\includegraphics[scale=\picScaleCb]{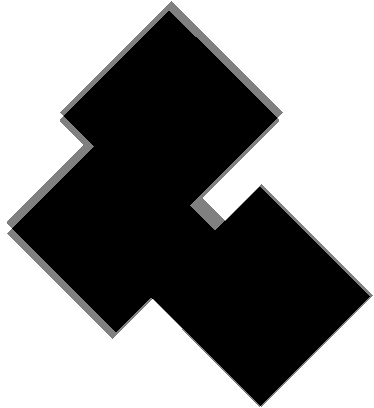}&
		\includegraphics[scale=\picScaleCb]{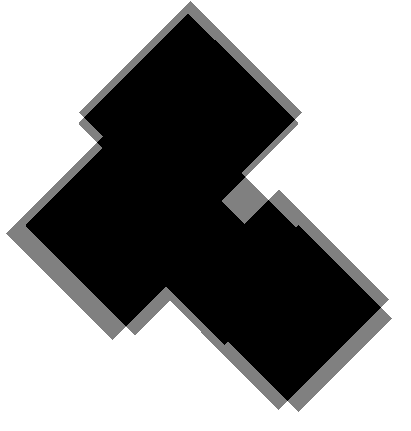}\\
		\includegraphics[scale=0.069]{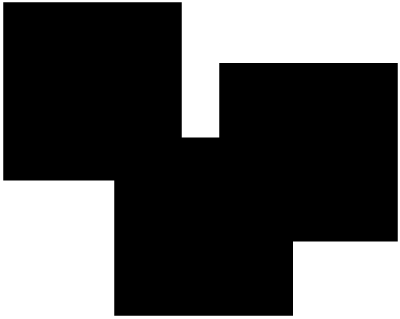}&
		\includegraphics[scale=\picScaleCc]{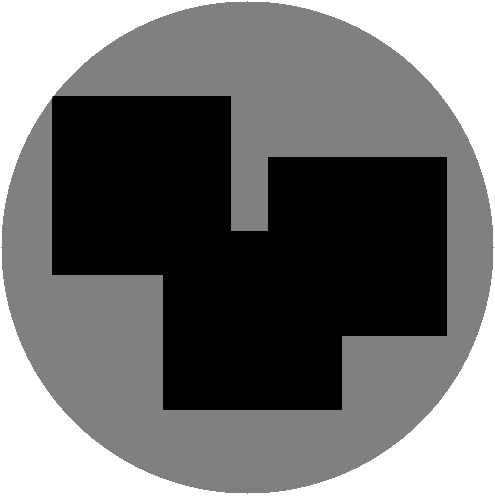}&
		\includegraphics[scale=\picScaleCc]{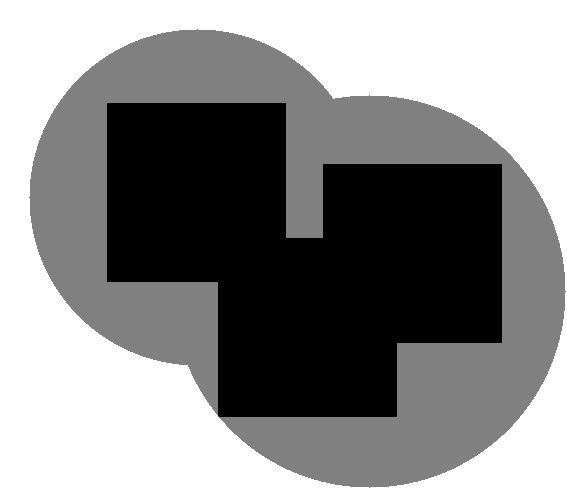}&
		\includegraphics[scale=\picScaleCc]{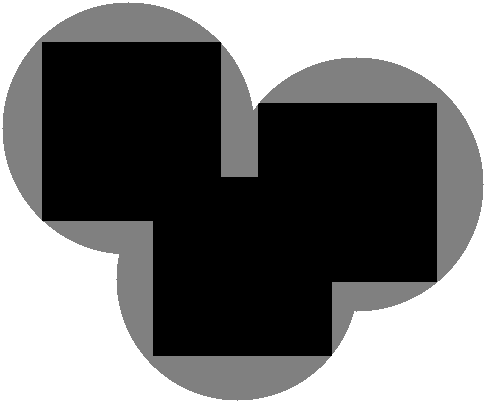}&
		\includegraphics[scale=\picScaleCc]{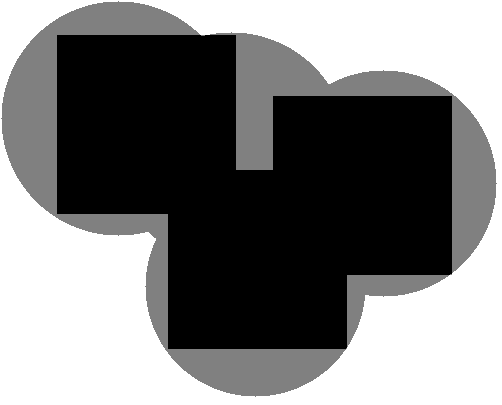}&
		\includegraphics[scale=\picScaleCc]{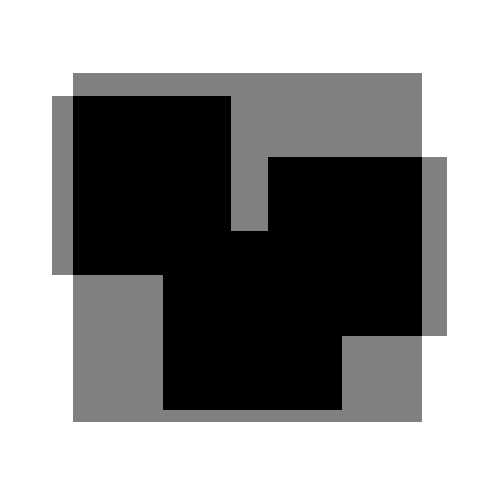}&
		\includegraphics[scale=\picScaleCc]{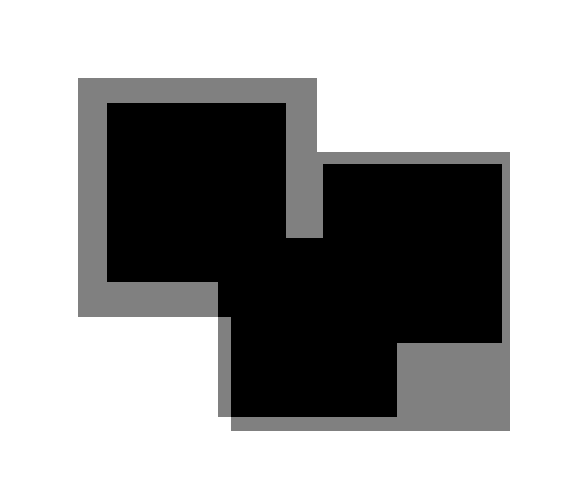}&
		\includegraphics[scale=\picScaleCc]{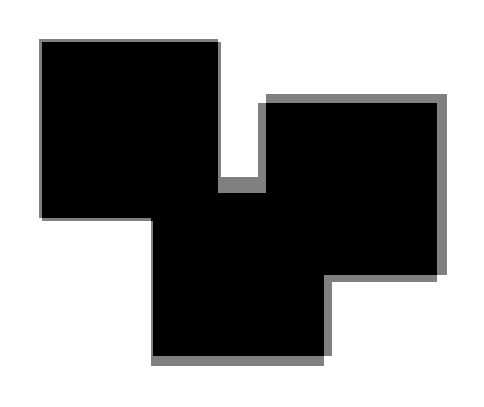}&
		\includegraphics[scale=\picScaleCc]{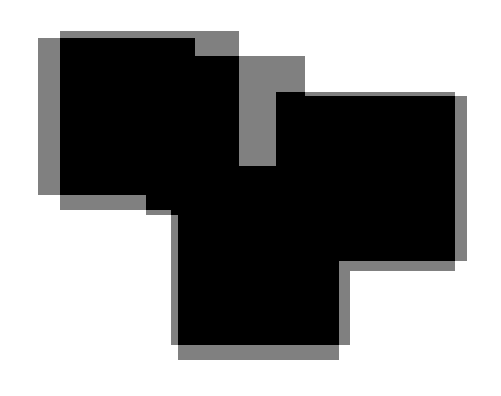}&
		\includegraphics[scale=\picScaleCc]{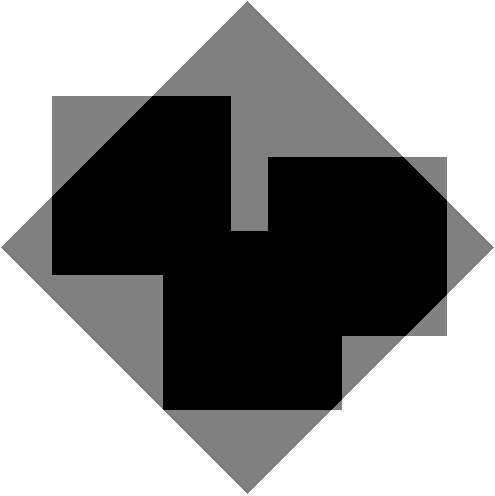}&
		\includegraphics[scale=\picScaleCc]{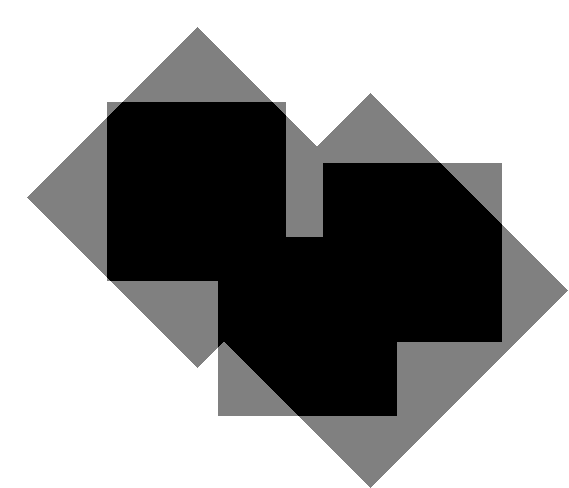}&
		\includegraphics[scale=\picScaleCc]{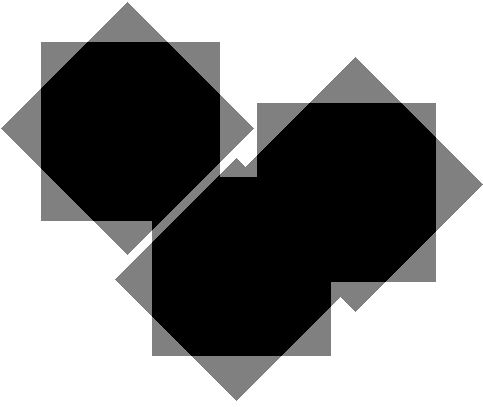}&
		\includegraphics[scale=\picScaleCc]{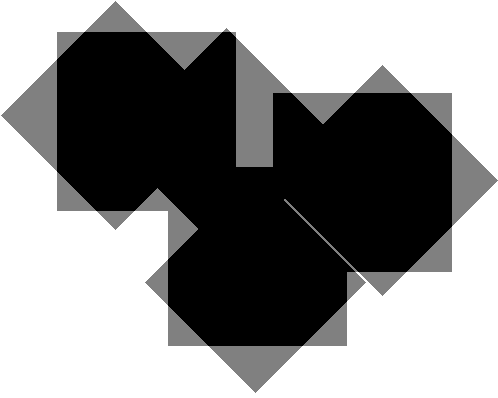}
	\end{tabular}
	\caption{Examples of re-visualization with different cluster numbers and marker types (circle, square, and diamond). Gray areas denote the symmetric difference between re-visualization and corresponding inputs. \label{fig_revis_corr}}
\end{figure}

For clustering algorithms, we use the typical distance-based K-Means, which partitions the data space into Voronoi cells through minimizing within-cluster variances namely squared Euclidean distance. Figure \ref{fig_revis_corr} shows examples of the re-visualization on in the clustering results with three common marker types namely circle, square and diamond.

\begin{figure}[h]
	\centering
	\small
	\setlength{\tabcolsep}{0.008\columnwidth}
	\begin{tabular}{ccc}
			\includegraphics[width=\picScaleA\columnwidth]{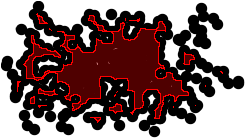}&
			\includegraphics[width=\picScaleA\columnwidth]{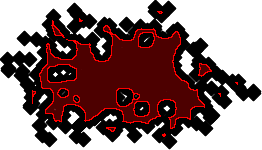}&
			\includegraphics[width=\picScaleA\columnwidth]{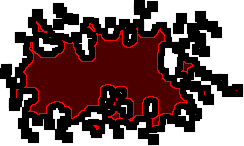}
		\end{tabular}
	\caption{Examples of severely-overlapping and completely-overlapping marks in scatter images. The red regions in images denote the areas where it may be impossible to recover how many and where data points are underlying.\label{fig_impossible_regions}}
\end{figure}

A crucial question in this way is whether re-visualization on clusters and original visualization have one-to-one correspondence. The answer is "No". There are two main reasons. On the one hand, the problem is ill-posed as it does not have a unique solution for severely-overlapping marks even for human vision.
For example, it may remain unknown about how many data points hiding in the red mask regions in Figure \ref{fig_impossible_regions} for ever if the graph maker does not disclose the ground truth. 
So, our aim in this work is only to find the minimum number of marks forming scatter images. On the other hand, re-visualizing marks on clustering results is affected by many factors: the centroid of a cluster may not be the real geometric centre of the original mark; radii calculated on clusters are affected by rounding accuracy; clustering results can not completely coincide with the original marks.

\subsection{Formulation} \label{sec:formulation}
How to estimate the number of clusters and the original marker type is crucial for this \textit{re-visualization on clustering} way. Here, we formulate it as an optimization problem.

Formally, given scatter image $I$, for a connected region $\mathbb{C}\subset I$, the problem to be solved can be formulated as 
\begin{equation}
	\begin{array}{ll}
		\min & {f(\mathbb{C},n,m)}\\
		s.t. & 1 \leqslant n \leqslant N_0\\
		& m \in \mathbb{S}
	\end{array},
	\label{eq_multiOpt}
\end{equation}
where the variables $n$, $m$ denote the number of clusters and the type of markers respectively, and $N_0$ denotes the possibly maximum numbers of clusters, and $\mathbb{S}$ is a finite set containing supporting markers. Considering the uncertain scale of input images, the value of $N_0$ should be associated with $\mathbb{C}$. It can be set as 
\begin{equation}
	\begin{array}{lr}
		N_0=\frac{|\mathbb{C}|}{\mathfrak{F}}, & 1 < \mathfrak{F} < |\mathbb{C}|,
	\end{array}
	\label{eq_factorF}
\end{equation}
where $\mathfrak{F}$ is a space setting factor. The factor $\mathfrak{F}$ is used to set the range of search space for the number of cluster $n$. A greater $\mathfrak{F}$ is helpful to reduce the search space but may miss the true solution, and a smaller $\mathfrak{F}$ will increase the search space but will increase the time consumption to find the optimal solution.

\begin{figure}[h]
	\centering
	\includegraphics[width=\columnwidth]{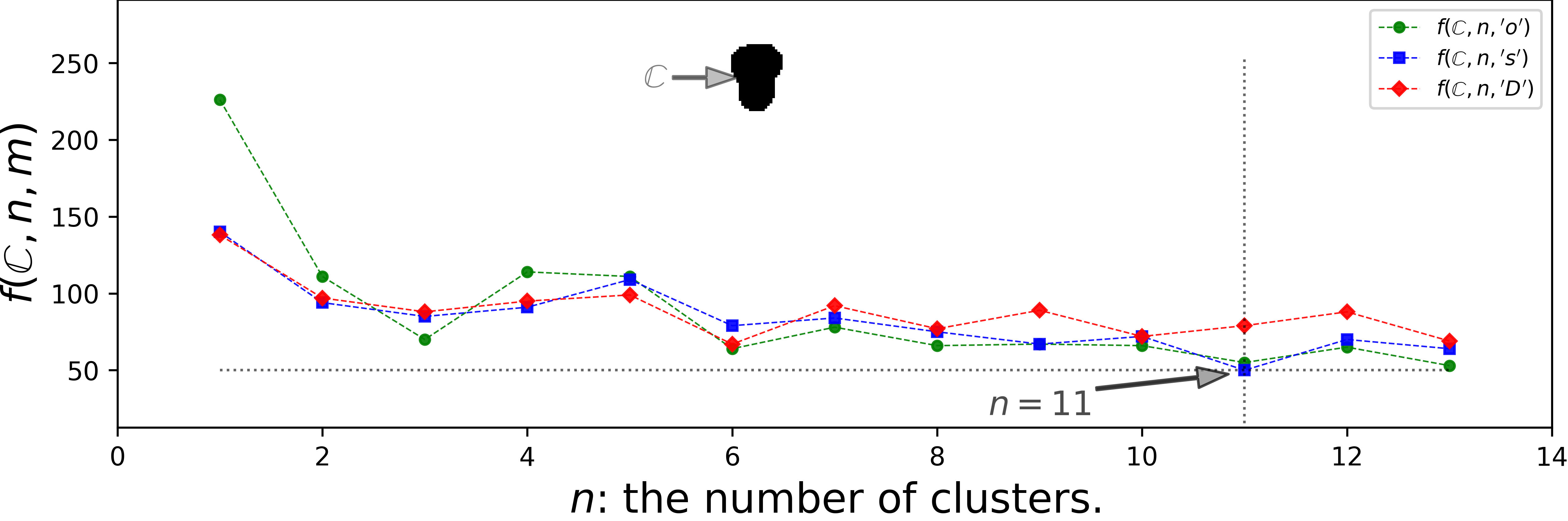}
	\caption{An example to show the variation of the first objective $f(\mathbb{C},n,m)$ with increasing the number of clusters $n$ and changing the marker type. $\mathbb{C}$ denotes the input connected region.\label{fig:curve_of_diff}}
\end{figure}

Regarding how to measure the difference between the original visualization and its re-visualization. We use the symmetric difference to measure such difference. Formally, for a connected region $\mathbb{C}$ and its re-visualization $\mathbb{V}$, the objective function can be expressed as 
\begin{equation}
	\begin{array}{rcl}
		f(\mathbb{C},n,m) & = & |\mathbb{C} \triangle \mathbb{V}| \\
		\mathbb{V} & = & revis(km(\mathbb{C},n),m)
	\end{array},
	\label{eqSymmDiff}
\end{equation}
where $km(\mathbb{C},n)$ means using K-means algorithm to divide $\mathbb{C}$ into $n$ groups. $revis$ function uses the clustering results and the marker type $m$ to re-visualize marks in a new binary image.
The difference between original region $\mathbb{C}$ and its re-visualization $\mathbb{V}$ is represented by the cardinal number of the pixel-wise symmetric difference between $\mathbb{C}$ and $\mathbb{V}$. Figure \ref{fig:curve_of_diff} shows how the difference changes when adjusting the number of clusters and the marker type. But, the marker type and the number of clusters are wrong when the objective $f(\mathbb{C},n,m)$ in Eq. \ref{eqSymmDiff} reaches a local minimum.

Why can't the minimization of Eq. \ref{eqSymmDiff} reflect the right division of overlapping marks? Because the difference between a connected region and its re-visualization on clusters will gradually decrease with the increase of the number of clusters. The most extreme case is that the difference will become 0 when the number of clusters is equal to the number of pixels in the connected region. As the example shown in Figure \ref{fig:curve_of_diff}, the correct number of clusters should be 3, but the number of clusters is 11 wrong when $f(\mathbb{C},n,m)$ reach a global minimum. So, we are introducing other two prior constraints.

As mentioned in subsection \ref{sec:revis_on_clst}, our aim is to find the minimum number of marks to form scatter images. We can achieve this by imposing a prior constraint on the number of marks. In other words, we should keep the number of clusters as less as possible on the basis of satisfying the objective defined in Eq. \ref{eqSymmDiff}. It can be express as a function like
\begin{equation}
	g(n) = \frac{n}{N_0}f(\mathbb{C},n,m) + n\sqrt{\mathfrak{F}}. \label{eqGn}
\end{equation}
where $\frac{n}{N_0}$ can be treated as a proportion factor to the main objective $f(\mathbb{C},n,m)$. Adding the item $n\sqrt{\mathfrak{F}}$ in $g(n)$ is also important, because it can adjust the effect of $n$ directly and associated with the space setting factor $\mathfrak{F}$. The effect of constraint $g(n)$ is to make the number of marks as less as possible. It can balance out the effect of gradual failure of the main objective function $f(\mathbb{C},n,m)$ with increasing the number of clusters, and it makes re-visualization on clusters and original visualization have one-to-one correspondence on almost all cases.

Furthermore, considering that the marks in many scatter images have the same size, we introduce another optional prior constraint $h(\mathbb{C},n)$ to enforce the radii of all clusters to be as close as possible. The standard deviation can be used to express this constraint, namely 
\begin{equation}
	h(\mathbb{C},n) = std(radius(km(\mathbb{C},n))).  \label{eqHcn}
\end{equation}
Notably, $h(\mathbb{C},n)$ usually affected by the randomness of K-means clustering results especially when $n$ becomes large. The role of $h(\mathbb{C},n)$ is only used to fine-tine the results when all overlapping marks in a connected region have a same size. If the marks in a scatter image do not have a same size, this objective should be ignored.

Comprehensively considering the main objective function and two prior constraints, we can finally formulate the whole problem as a multi-objective optimization problem as 
\begin{equation}
	\begin{array}{ll}
		\min & {f(\mathbb{C},n,m), g(n), h(\mathbb{C},n)}\\
		s.t. & 1 \leqslant n \leqslant N_0\\
		& m \in \mathbb{S}
	\end{array}.
	\label{eq_multiOpt}
\end{equation}
Obviously, this multi-objective optimization problem is difficult to solve directly. So, we can convert it to a single-objective optimization as 
\begin{equation}
	\begin{array}{lrl}
		\min && \mathcal{L}(\mathbb{C},n,m) \\
		     &=& f(\mathbb{C},n,m)+\alpha g(n)+\beta h(\mathbb{C},n)\\
		s.t. && 1 \leqslant n \leqslant N_0\\
		&& m \in \mathbb{S}
	\end{array}.
	\label{eq_singleOpt}
\end{equation}
where $\alpha$, $\beta$ are the weights to control the influence of two prior constraints respectively. For the setting of $\alpha$, it is basically stable and weakly affected by the input scatter images, and it works well when setting $\alpha$ to a fixed real number around $1.0$. For the setting of $\beta$,  it depends on the characteristics of scatter images. For examples, if the size of marks in a scatter images are nearly same, it is better to set $\beta$ to a real number larger than zero, otherwise it is better to set it to zero.

\begin{figure}[h]
	\centering
	\includegraphics[width=\columnwidth]{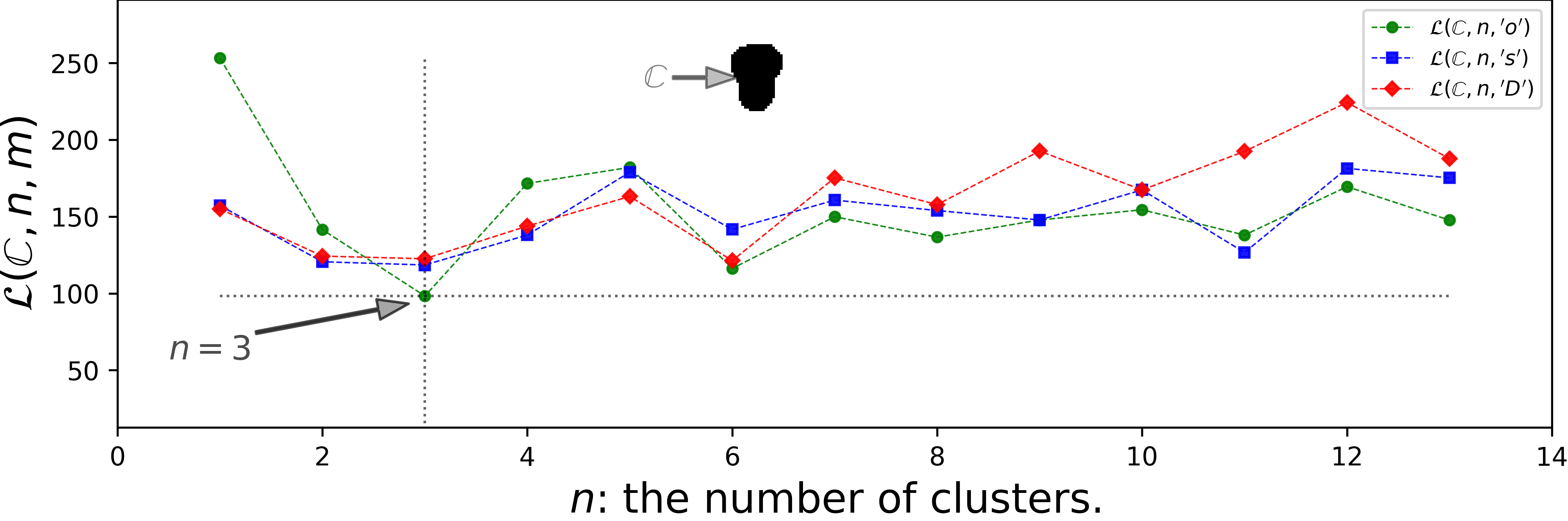}
	\caption{The variation of the objective function $\mathcal{L}(\mathbb{C},n,m)$ when $\alpha=1.5$ and $\beta=5.0$ with the number of clusters for the case in Figure \ref{fig:curve_of_diff}. \label{fig:curve_of_loss}}
\end{figure}

Take the Equations \ref{eqSymmDiff}, \ref{eqGn} and \ref{eqHcn} into this single object function $\mathcal{L}$, it can be expressed as

\begin{equation}
	\begin{array}{ll}
		\mathcal{L}(\mathbb{C},n,m) & = f(\mathbb{C},n,m) + \alpha (\frac{n}{N_0}f(\mathbb{C},n,m)+ n\sqrt{\mathfrak{F}}) + \beta h(\mathbb{C},n)\\
		& = (1+\alpha \frac{n}{N_0})|\mathbb{C} \triangle revis(km(\mathbb{C},n),m)| + \alpha n\sqrt{\mathfrak{F}} + \beta std(radius(km(\mathbb{C},n)))
	\end{array},
	\label{eq_singleOptD}
\end{equation}
For the example in Figure \ref{fig:curve_of_diff}, its curve of loss function $\mathcal{L}$ with $\alpha=1.5$ and $\beta=5.0$ is drawn in Figure \ref{fig:curve_of_loss}. It is obviously shown that the minimization of the single objective $\mathcal{L}$ becomes the global minimization at $n=3$ and marker type is circle. It indicates that it is feasible to formulate the mark localization problem as a single-objective optimization problem defined in Eq. \ref{eq_singleOpt}.

Finally, the number of clusters and marker type can be solved as
\begin{equation}
	\hat{n}, \hat{m} = \argmin_{n, m}{\mathcal{L}(\mathbb{C},n,m)} \label{eq_argmin},
\end{equation}
where $\hat{n}$, $\hat{m}$ are the optimal solution when the objective function $\mathcal{L}$ reaches the global minimum.

\subsection{Optimization}
Obviously, the above single objective optimization problem defined in Eq. \ref{eq_argmin} is not a convex optimization problem. In this work, we employ Simulated Annealing (SA), a Monte Carlo search method, to solve this optimization. SA is a widely-used probabilistic technique for approximating the global optimum of a given function of many independent variables, and often used in combinatorial or non-convex optimization problems \cite{INGBER_199329,ScottKirkpatrick1984} such as the travelling salesman problem \cite{kirkpatrick_optimization_1983}.

\begin{algorithm}[!h]
	\DontPrintSemicolon
	\caption{Pseudo-code of optimizing process.}\label{alg:sa}
	\KwIn{$\mathbb{C}$       \tcp*[r]{A connected region.}\
		\hspace*{10.5mm} $\mathbb{S}$    \tcp*[r]{The set of supportable markers.}
		\hspace*{10.5mm} $\gamma_s$     \tcp*[r]{Coefficient of stop criteria.}
		\hspace*{10.5mm} $\gamma_m$       \tcp*[r]{Coefficient of Markov steps.}
	}
	\KwOut{ $\hat{n}, \hat{m}$   \tcp*[r]{Approximated optimum.}}
	$N_0 \gets \max(1, |\mathbb{C}|/\mathfrak{F})$\;
	$T_0 \gets N_0$\;
	$S_s \gets \gamma_s * \log{(N_0*|\mathbb{S}|)} $  \tcp*[r]{The number of steps for stopping.}
	$S_m \gets \gamma_m * \log{(N_0*|\mathbb{S}|)} $  \tcp*[r]{The number of Markov steps.}
	$\ell \gets \min(\mathcal{L}(\mathbb{C},1,m_i)), m_i \in \mathbb{S}$\;
	$n,m \gets \argmin(\mathcal{L}(\mathbb{C},1,m_i)), m_i \in \mathbb{S}$\;
	$\dot{\ell}, \hat{\ell},\hat{n}, \hat{m} \gets \ell,\ell,n,m $\;
	$b_c \gets 0$  \tcp*[r]{Counter for best solutions.}
	$t_c \gets 0$  \tcp*[r]{Counter for temperature changes.}
	\While{$T > 0.1 \textbf{ and } b_c < S_s$}{
		\For{$i \gets 1$ \KwTo $S_m$}{
			$\tilde{n},\tilde{m} \gets news(n,m,N_0,\mathbb{S})$  \tcp*[r]{New solution.}
			$\tilde{\ell} \gets \mathcal{L}(\mathbb{C},\tilde{n},\tilde{m})$\;
			$p \gets min(1, \exp(-(\tilde{\ell}-\ell)/T))$\;
			$r \sim U(0,1)$\;
			\If{$\tilde{\ell}<\hat{\ell}$}{
				$\hat{n}, \hat{m}, \hat{\ell} \gets \tilde{n}, \tilde{m}, \tilde{\ell}$\;
			}
			\If{$\tilde{\ell}<\ell$ \textbf{ or } $p>r$ }{
				$n,m,\ell \gets \tilde{n}, \tilde{m},\tilde{\ell}$ 
				 \tcp*[r]{Accept.}
			}
		}
		\eIf{$\hat{\ell} < \dot{\ell}$}{
			$b_c \gets 0$			
		}{
			$b_c \gets b_c+1$
		}		
		$T \gets T_0/(1+\log(1+t_c))$ \tcp*[r]{Reduce the temperature.}
		$t_c \gets t_c+1$\;
		$\dot{\ell} \gets \hat{\ell}$\;
	}
\end{algorithm}

Following the basic framework of SA in \cite{Emaile1989}, we design a variant of SA algorithm especially for this optimization problem. Its pseudo-code is presented in Algorithm \ref{alg:sa}. 
In this algorithm, we use the generalized Metropolis acceptance criterion $min(1, \exp(-(\tilde{\ell}-\ell)/T))$ \cite{metropolis_equation_1953} to decide whether accept a solution in a certain probability. A proper neighbourhood structure is crucial for SA algorithm. The function $news$ is a neighbour function to randomly generate new candidate solutions for variables based on the current solution. Here, for accelerating the optimization process, we assume the variation of $m$ and $n$ for a Markov step subject to two-dimensional Gaussian distribution, namely
\begin{equation}
	\begin{bmatrix}
	\Delta{n} & \Delta{m}
	\end{bmatrix}^T = \Delta{z} \sim \mathcal{N}(0, \Sigma)
\end{equation}
where $\Sigma$ is the covariance matrix. Considering dramatical variation of the scale of solution spaces for different binary regions and the less correlation between two variables, we set the covariance matrix $\Sigma$ adaptively as 
\begin{equation}
	\Sigma = \begin{bmatrix}
		(\frac{N_0}{c_\sigma})^2&0\\
		0&(\frac{|\mathbb{S}|}{c_\sigma})^2
	\end{bmatrix}
\end{equation}
according to the scale of space $N_0$ and $|\mathbb{S}|$. The parameter $c_\sigma$ controls the range of neighbourhood, which equals to a constant 6 in this work.
Then, a new solution $(\tilde{n}, \tilde{m})$ can be calculated by 
\begin{equation}
	\tilde{n} = int(n+\Delta{n}) \% N_0
\end{equation} and 
\begin{equation}
	\tilde{m} = int(m+\Delta{m}) \% |\mathbb{S}|
\end{equation}
where $(n,m)$ denotes the current solution.

As the normal structure of a SA algorithm, we also take two-layers loop program to control the optimization process. The outer loop controls the termination conditions through temperature and solution stability, and the inner loop finishes a Markov process under a certain temperature. As same as in common SA, there are still two intractable problems, namely when to terminate the outer loop and how many steps in the inner loop. Another challenge in this work is the long span of different scales of connected regions, which brings about a consequential difficulty that same stop criterion and same Markov steps may be infeasible. Therefore, we propose an adaptive control method to fit different connected regions. Given $N_0$ and $\mathbb{S}$, the number of Markov step $S_m$ is 
\begin{equation}
	S_m = \gamma_m*log(N_0*|\mathbb{S}|) \label{eqGammaM}
\end{equation}
and the stop criteria is 
\begin{equation}
	b_c < S_s = \gamma_s*log(N_0*|\mathbb{S}|) \label{eqGammaS}
\end{equation}
where $b_c$ is a counter to monitor how many times the best solution in an inner loop has not changed with the decrease of temperature. This stop criteria aims to find an optimal solution which is stable enough. $\gamma_m$ and $\gamma_s$ are parameters to control the solving process of simulated annealing. How to choose good values for $\gamma_m$ and $\gamma_s$ is a compromise between the optimization speed and accuracy. Totally, $\gamma_m$ and $\gamma_s$ are very stable for different binary regions because the logistic function reflects the characteristic of SA for different input scales. Subsection \ref{sec:infl_sa} will discus this with experimental results.

\subsection{ Recognizing single marks in advance} \label{sec:rsma}
In most scatter images, there are some isolated single marks which are usually detected more easily. So, an improved approach is to utilize single marks to recognize the type and size of marker in advance. We call it RSMA (Recognizing Single Marks in Advance). RSMA is helpful to define an automatically self-adaptive search space more appropriately. ScatterScanner \cite{baucom_scatterscanner_2013} ever used a similar idea to obtain the size of Gaussian kernel.

The value of $\mathfrak{F}$ in Eq. \ref{eq_factorF} can be dynamically calculated by using RSMA. Given the size of single marks, $\mathfrak{F}$ can be determined as
\begin{equation}
	\begin{array}{lr}
		\mathfrak{F} = \kappa E(S), & \frac{1}{E(S)} < \kappa < \frac{|\mathbb{C}|}{E(S)}
	\end{array},
	\label{eqrsmia}
\end{equation}
where $S$ is a random variable denotes the size of single marks, and $\kappa$ is a control parameter to adjust the impact of expectation $E({S})$. Empirically $\kappa$ is better to set a little bit smaller than the overlapping severity.

\section{Dataset and metric} \label{subsec_dataset}
\subsection{Current datasets}
A proper dataset is crucial to evaluate the performance of various methods for a challenging task. For scatter marks localization task, some available public datasets such as FigureQA \cite{FigureQA2017}, PlotQA \cite{PlotQA2020wacv} and UB-PMC dataset \cite{CHARTInfo2022} may be considered as options. But they are not appropriate to be the benchmark datasets after deliberateness. For one thing, almost of all scatter images in these datasets contain many irrelevant elements for evaluating marks localization task, such as axes, ticks, labels.
For another, as far as we know, there are no datasets especially for evaluating overlapping marks localization and observing the relationship between performance and overlapping severity.

\subsection{A new dataset}
Therefore, we built a new dataset called SML2023 specially for evaluating the scatter mark localization task. This dataset contains 297 synthetic scatter images with different overlapping-severity. These scatter images were generated using data points with different quantity scale and various markers distinct in type and size. Every scatter image only contains the plot area to eliminate effects of other elements for evaluating mark localization. The SML2023 dataset could be found in https://github.com/ccqym/OsmLocator/datasets.

\begin{figure}[!h]
	\centering
	\small
	\setlength{\tabcolsep}{0.008\columnwidth}
	\begin{tabular}{ccccc}
		\includegraphics[width=\picScaleF\columnwidth]{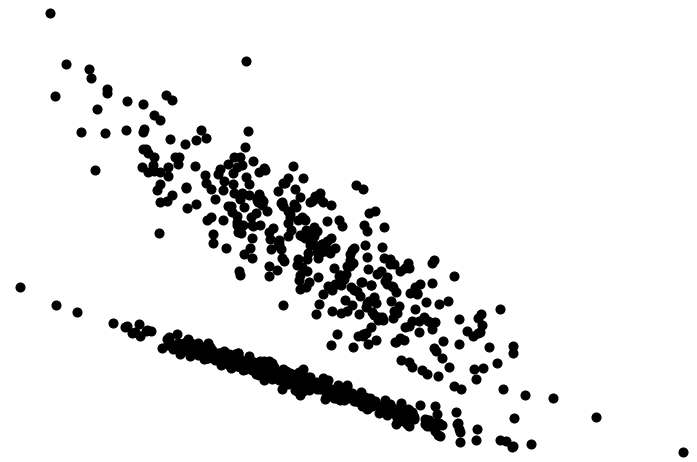}&
		\includegraphics[width=\picScaleF\columnwidth]{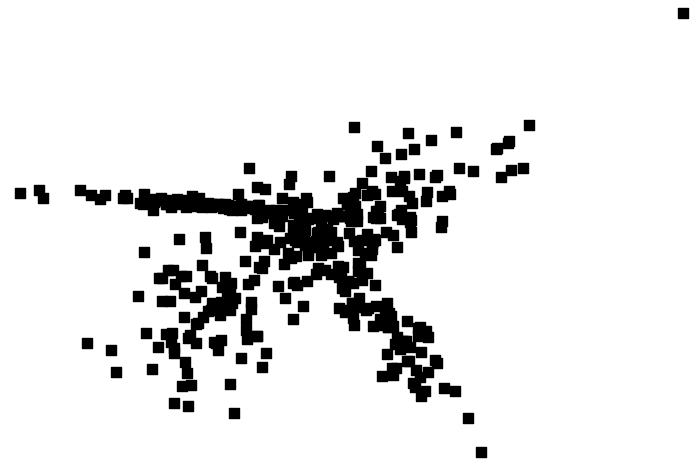}&
		\includegraphics[width=\picScaleF\columnwidth]{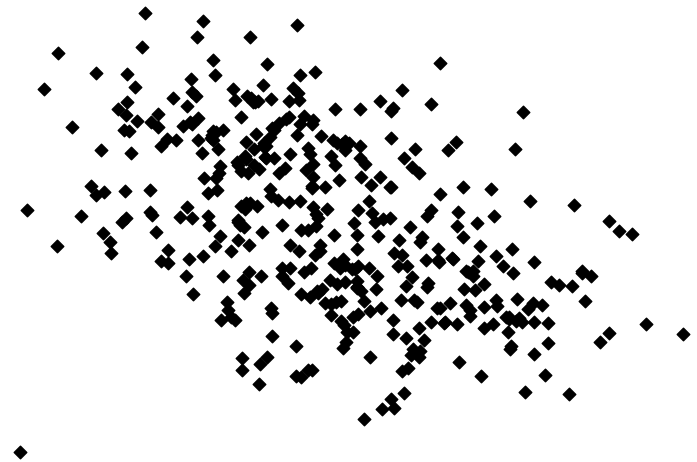}&
		\includegraphics[width=\picScaleF\columnwidth]{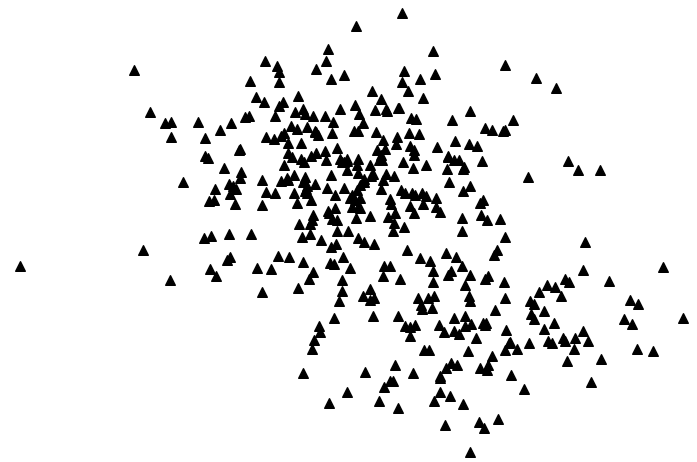}&
		\includegraphics[width=\picScaleF\columnwidth]{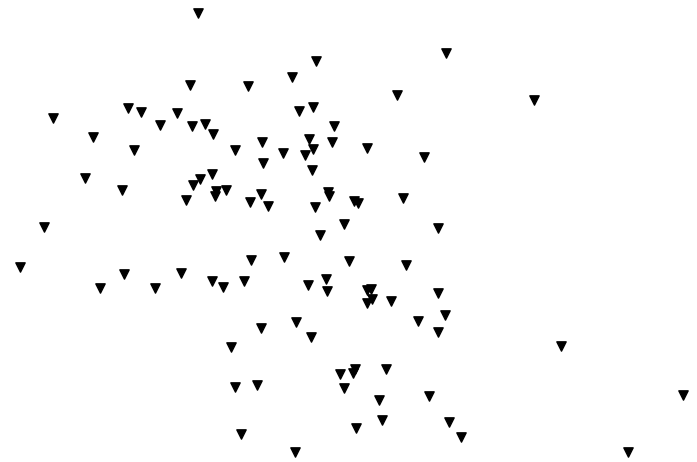}\\
		s=0.504&s=0.374&s=0.164&s=0.124&s=0.026\\
		\includegraphics[width=\picScaleF\columnwidth]{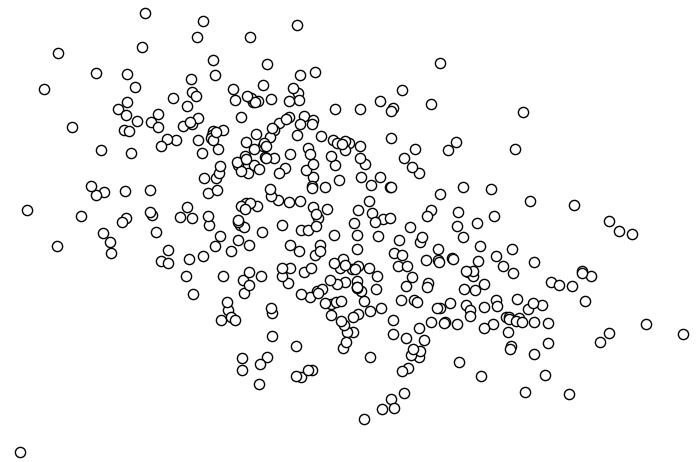}&
		\includegraphics[width=\picScaleF\columnwidth]{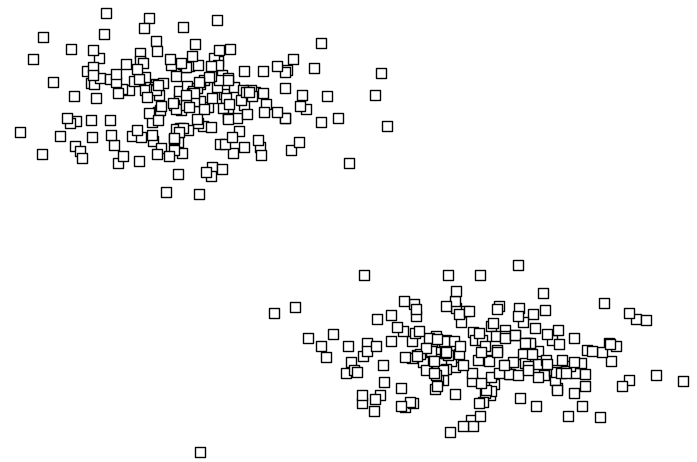}&
		\includegraphics[width=\picScaleF\columnwidth]{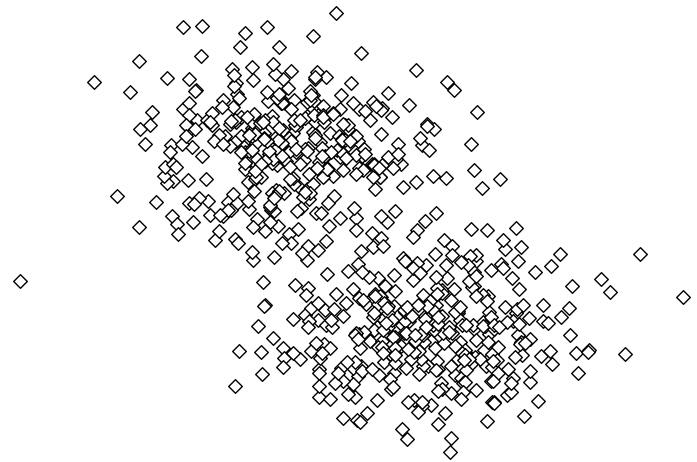}&
		\includegraphics[width=\picScaleF\columnwidth]{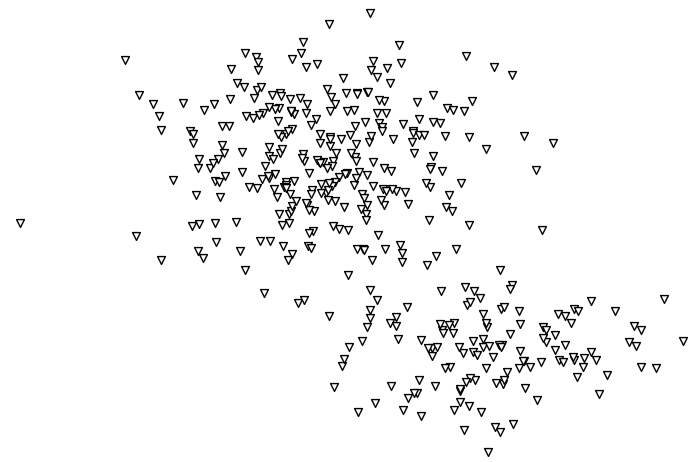}&
		\includegraphics[width=\picScaleF\columnwidth]{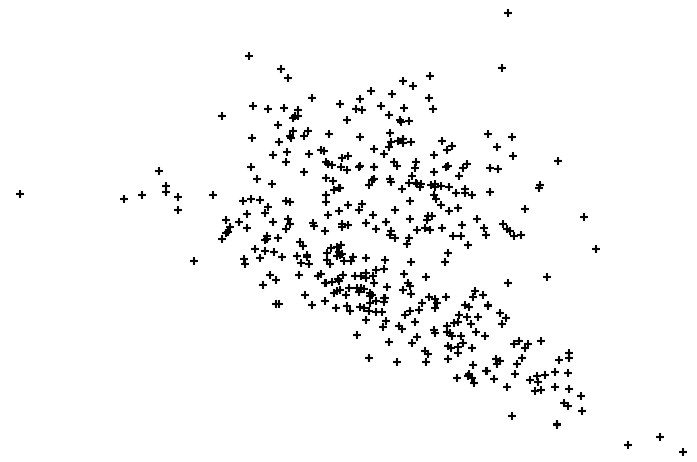}\\
		s=0.146&s=0.295&s=0.319&s=0.096&s=0.075\\
	\end{tabular}
	\caption{Examples in the synthetic dataset SML2023. The label below each scatter image indicts the corresponding overlapping severity of marks. \label{fig:dataset_examples}}
\end{figure}

There are 11 types of markers including $\bullet$, $\circ$, $\blacksquare$, $\square$, $\blacklozenge$, $ \lozenge$, $\blacktriangle$, $\vartriangle$, $\blacktriangledown$, $\triangledown$, $+$ in the dataset, and 27 scatter images for each marker type. The number of marks in a synthetic image is one of the set {100,400,700}. For a fixed number of marks and a marker type, 6 images were generated by isotropic Gaussian blobs with different parameters, and 3 images were generated by normally distributed hypercube for classification with 2 features.
Some examples of scatter images in the new dataset are shown in Fig. \ref{fig:dataset_examples}.

\subsection{Overlapping severity}
\begin{figure}[!h]
	\centering
	\includegraphics[width=0.8\columnwidth]{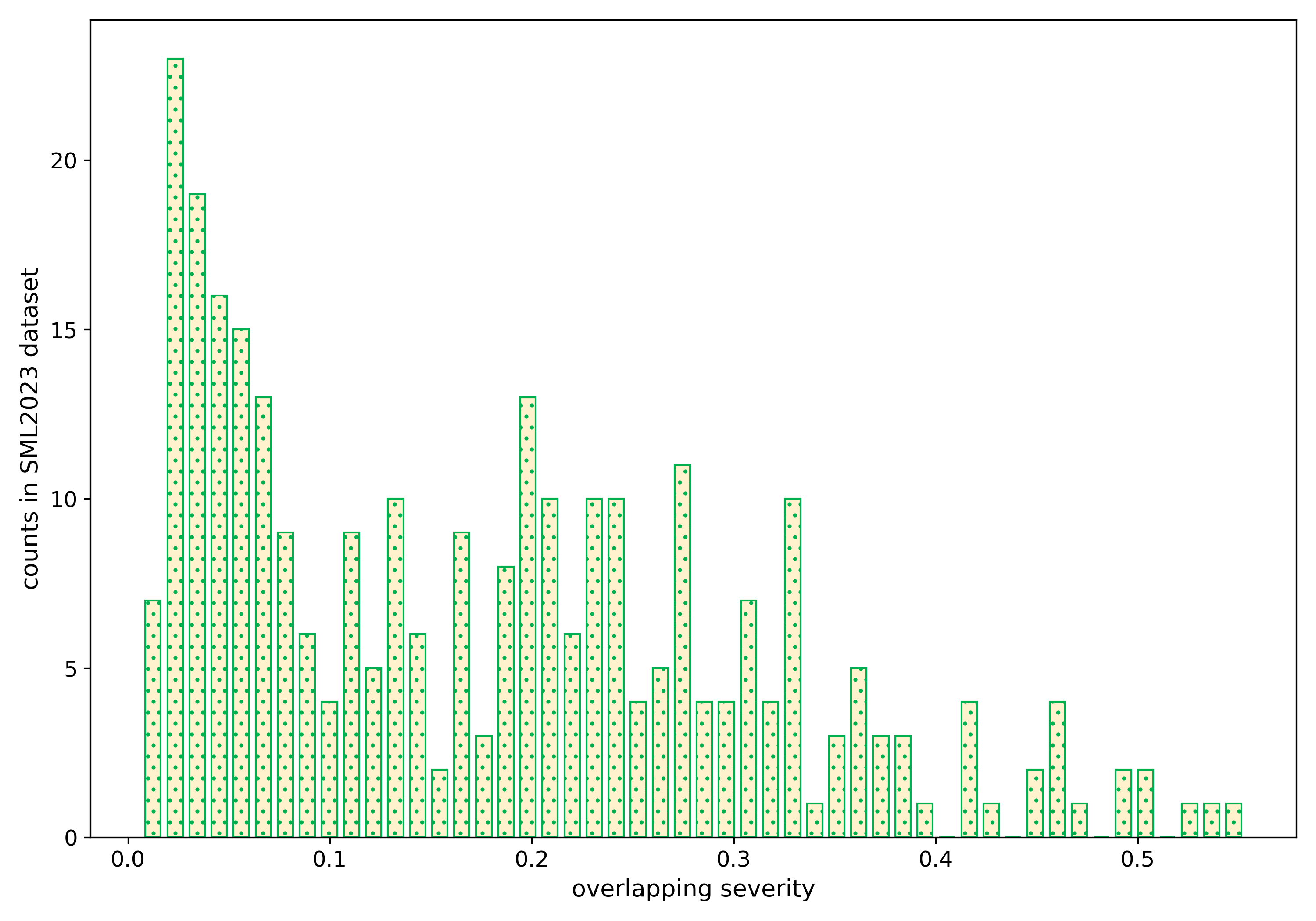}
	\caption{Counts in the synthetic dataset SML2023. \label{fig:dataset_counts}}
\end{figure}

For each scatter image, we annotated important information including overlapping severity, marks count and mark locations. The overlapping severity reflects the degree of overall overlapping of all marks and is calculated by the formula
\begin{equation}
	s = 1 - \dfrac{|\mathbb{O}|}{\sum_{i=1}^{Q}|\mathbb{M}_i|},
\end{equation}
in which $\mathbb{O}$ is the set of all mark pixels in a scatter image, and $\mathbb{M}_i$ is the set of all pixels of one mark separately in all marks $Q$. Fig. \ref{fig:dataset_counts} show the histogram of overlapping severities of all scatter images. The label below each scatter image in Fig. \ref{fig:dataset_examples} indicts the corresponding overlapping severity of marks.

Notably, only one marker type was used to draw every scatter image in the dataset, namely no overlapping of different markers in one scatter image. Because different markers usually be distinguished by different colours. There are seldom scatter images in which different markers use a same colour like in the paper \cite{browuer_segregating_2008}. In practice, it is better to extract connected regions for different markers by colour information and then use the proposed method to locate marks for each marker.

\subsection{Metric} \label{sec:metrics}
We used the metric in CHART-Info competitions \cite{chart_info_2023} to evaluate mark localization. It treats the evaluation as a job-assignment problem to find the minimum cost to pair each GT point with a localized point. 
Formally, given a set of localized marks $\mathbb{P}$ and a set of marks to the ground truth $\mathbb{G}$, a capped distance between each pair $(p_i, g_i)$ is firstly calculated as 
\begin{equation}
	d(p_i, g_i) = min(1, \frac{\sqrt{(p_i-g_i)V^{-1}(p_i-g_i)}}{\lambda})
	\label{eqMetricLmd}
\end{equation}
where $V^{-1}$ is the inverse of covariance matrix of $\mathbb{G}$, and $\lambda$ is a hyper-parameter to adjust how far a pair away will be counted as effective. The distance of a pair farther than a standard deviation of $\mathbb{G}$ will be assigned the maximum distance 1 when $\lambda=1$. Then, a pairwise square cost matrix $C$ in which the value of each element $c_{i,j}$ is equal to the pair distance $d_{i,j}$, and the padded elements are assigned $1$.   
Next, the minimum pairing cost $c$ between $\mathbb{G}$ and $\mathbb{P}$ is resolved by taken it as a linear sum assignment problem.
\begin{equation}
	cost(\mathbb{G},\mathbb{P}) = \sum_{(i,j)\in \textbf{A}}c_{i,j}
\end{equation}
where (i,j) denotes a pair in the optimal assignment $\textbf{A}$.
Lastly, the final score is calculated as 
\begin{equation}
	s = 1- \frac{cost(\mathbb{G},\mathbb{P})}{max(|\mathbb{G}|,|\mathbb{P}|)}.
\end{equation}
We use it as our evaluation metric and call it assignment-cost-based metric \emph{ACB}. The overall metric can be taken the mean and the standard deviation of all \emph{ACB} scores.
To observe the influence of $\lambda$, we chose three $\lambda$ values $\{1,5,10\}$ to evaluate all cases.

\section{Experiments} \label{sec_expr}
\subsection{Implementation}
We developed a software tool named OsmLocator to implement the proposed method with Python, and tested its performance on the SML2023 dataset \ref{subsec_dataset}.
OsmLocator supports 11 kinds of common markers including $\bullet$,$\circ$,$\blacksquare$,$\square$,$\blacklozenge$,$ \lozenge$,$\blacktriangle$,$\vartriangle$,$\blacktriangledown$,$\triangledown$,$+$. Meanwhile, we also implemented the filter-based method ScatterScanner \cite{baucom_scatterscanner_2013} with python to compare performance.

We also considered to compare to a newest curvature based method \cite{ZouTong2021} , but it is infeasible to run on the scatter images in an acceptable time because its complexity is exponential and there are many connected regions consisting of too many overlapping marks.

We did not test deep-learning-base methods in our experiments, such as Scatteract \cite{cliche_scatteract_2017} and Yolo \cite{YoloV1_Redmon2016}. One the one side, all deep-learning-base methods are supervised, so it is unfair to compare to unsupervised methods. One the other side, references \cite{cliche_scatteract_2017,Tausif2023} had clearly stated that related deep-learning-based methods can not work well on overlapping objects.

With the implemented tool OsmLocator, we made some experiments in several aspects to test the performance and make clear how the control parameters affect the performance. 
The following subsections present the details and results of such experiments.

\subsection{Experiments without RSMA}
The space setting factor $\mathfrak{F}$ greatly affects the scale and structure of solution space, and should be adjusted according to the size of markers. So, it is interest to know how the space setting factor affect the performance. To this end, we fixed the control parameters that are roughly stable for different cases and ran OsmLocator on SML2023 dataset with different $\mathfrak{F}$ values to observe the changes of performance.

In details, we set the control parameters as $\alpha=1.1, \beta=1.0, \gamma_s=1.5, \gamma_m=1.5$ and assigned each number in the set {10,30,60,90,120,150} to $\mathfrak{F}$, and then tested OsmLocator with these parameters and recorded the corresponding ACB scores.
In order to compare the performance of ScatterScanner \cite{baucom_scatterscanner_2013} in same situation, we also ran ScatterScanner with four different kernel sizes namely {5,10,15,20} to observe the performance changes.

\begin{table}[!h]
	\begin{center}
		\setlength{\tabcolsep}{1.0em}
		{\caption{Experimental results without RSMA}\label{tb_without_rsma}}
		\begin{tabular}{lcccc}
			\hline
			Method & ACB[$\lambda=1$] & $ACB[\lambda=5$] & ACB[$\lambda=10$] & Time consumption\\
			\hline
			ScatterScanner[$k=5$]  & 0.274$\pm$0.267 & 0.285$\pm$0.279 & 0.287$\pm$0.281 & 1.437$\pm$0.143\\
			ScatterScanner[$k=10$]  & 0.471$\pm$0.249 & 0.481$\pm$0.255 & 0.483$\pm$0.257 & 1.334$\pm$0.14 \\
			ScatterScanner[$k=15$]  & 0.325$\pm$0.233 & 0.332$\pm$0.237 & 0.333$\pm$0.238 & 1.32$\pm$0.157 \\
			ScatterScanner[$k=20$]  & 0.168$\pm$0.186 & 0.171$\pm$0.187 & 0.172$\pm$0.187 & 1.304$\pm$0.084 \\
			OsmLocator[$\mathfrak{F}=10$]   & 0.779$\pm$0.150 & 0.797$\pm$0.154 & 0.800$\pm$0.154 & 198.4$\pm$333.2 \\
			OsmLocator[$\mathfrak{F}=30$]   & 0.778$\pm$0.149 & 0.797$\pm$0.153 & 0.800$\pm$0.154 & 196.3$\pm$353.4 \\
			OsmLocator[$\mathfrak{F}=60$]   & 0.778$\pm$0.147 & 0.797$\pm$0.150 & 0.799$\pm$0.150 & 190.3$\pm$323.9 \\
			OsmLocator[$\mathfrak{F}=90$]   & 0.780$\pm$0.147 & 0.799$\pm$0.150 & 0.801$\pm$0.151 & 210.8$\pm$373.4 \\
			OsmLocator[$\mathfrak{F}=120$]   & 0.777$\pm$0.150 & 0.795$\pm$0.153 & 0.797$\pm$0.153 & 195.4$\pm$337.3 \\
			OsmLocator[$\mathfrak{F}=150$]   & 0.778$\pm$0.151 & 0.798$\pm$0.154 & 0.800$\pm$0.155 & 195.8$\pm$369.0 \\
			\hline
			\multicolumn{5}{p{405pt}}{\textit{All experiments in this table were done with fixed control parameters $\alpha=1.1, \beta=1.0, \gamma_s=1.5, \gamma_m=1.5$. $\mathfrak{F}$ is the space setting factor defined in Eq. \ref{eq_factorF}. $k$ denotes the kernel size of ScatterScanner \cite{baucom_scatterscanner_2013}. $\lambda$ is the evaluation hyper-parameter in Equation \ref{eqMetricLmd}.}}
		\end{tabular}
	\end{center}
\end{table}

Table \ref{tb_without_rsma} lists the experimental results. For one thing, the proposed method gains 30.9\% improvement on ACB metric at best in comparison with ScatterScanner. For another, it is shown that the control parameter $\mathfrak{F}$ has similar performance in a wide range. This indicates that our formulation for scatter marks localization seize the essence of the problem and the space setting factor of the formulation has great tolerance for different values. In another word, our method is insensitive to the space setting factor, and this is very helpful in applications. 

For time consumption, our method spends extremely more time on average than ScatterScanner. It is because heavy overlapping cases spent extremely high time consumption cost and pulled up the average. We will presents the detailed analysis of time consumption in the next subsection \ref{sec:exprsm}.
Moreover, it is a little surprised that the mean time consumption of OsmLocatter on selected $\mathfrak{F}$ values are little difference. Generally, a greater $\mathfrak{F}$ generally means a smaller solution space and should brought less time consumption. But the number of iterations of simulated annealing in Algorithm \ref{alg:sa} is mainly affected by the Equations \ref{eqGammaM} and \ref{eqGammaS}, which are in logarithm relation with the size of a solution space. Therefore, with the increase of solution space, the time consumption changes less for a given connected regions. What does affect time consumption? The answer is the parameters of simulated annealing $\gamma_s$ and $\gamma_m$. The latter subsection \ref{sec:infl_sa} will discuses the time consumption of different $\gamma_s$ and $\gamma_m$ parameters. It's worth noting that not every case our method costed long time to finish, and actually in some light overlapping cases, our methods spent less time than ScatterScanner. We will presents the concrete analysis in the next subsection \ref{sec:exprsma}. 

Although the space setting factor $\mathfrak{F}$ for all cases in datasets has weak influence on the performance and time consumption as explained above, but $\mathfrak{F}$ indeed affects the performance for single cases, so a better choose is to use RSMA in the subsection \ref{sec:rsma}.

\subsection{Experiments with RSMA} \label{sec:exprsma}
As mentioned in subsection \ref{sec:rsma}, it is a possible way to detect isolated single marks to determine the value of the space setting factor $\mathfrak{F}$. In order to validate whether RSMA is feasible and compare to related works, we tested the performance of ScatterScanner \cite{baucom_scatterscanner_2013} and the proposed OsmLocator on SML2023 dataset with RSMA. Like the experiments in the previous subsection, these experiments with RSMA were done with same fixed control parameters $\alpha=1.1, \beta=1.0, \gamma_s=1.5, \gamma_m=1.5$.

\begin{table}[!h]
	\begin{center}
		\setlength{\tabcolsep}{1.0em}
		{\caption{Experimental results with RSMA}\label{tb_with_rsma}}
		\begin{tabular}{lcccc}
			\hline
			Method & ACB[$\lambda=1$] & ACB[$\lambda=5$] & ACB[$\lambda=10$]&Time consumption\\
			\hline
			ScatterScanner+RSMA  & 0.468$\pm$0.257 & 0.476$\pm$0.262 & 0.477$\pm$0.263 & 1.421$\pm$0.131 \\
			OsmLocator+RSMA & \textbf{0.779$\pm$0.153} &\textbf{ 0.798$\pm$0.157} & \textbf{0.801$\pm$0.158} & 95.5$\pm$182.7 \\
			\hline
			\multicolumn{5}{p{375pt}}{\textit{Notes are same with Table \ref{tb_without_rsma}.}}
		\end{tabular}
	\end{center}
\end{table}

\begin{figure}[!h]
	\centering
	\small
	\setlength{\tabcolsep}{0.008\columnwidth}
	\begin{tabular}{cc|cc}
		\includegraphics[width=\picScaleFr\columnwidth]{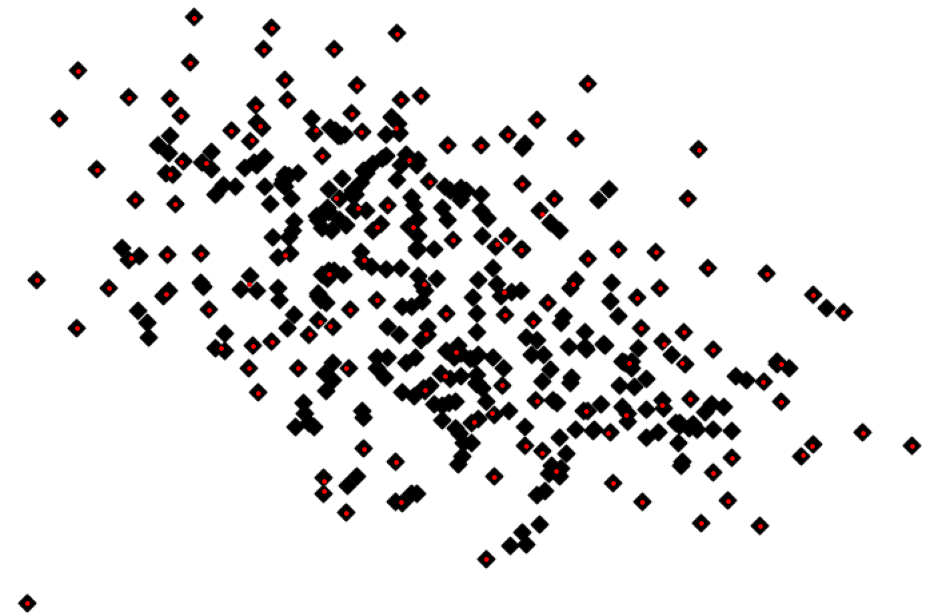}&
		\includegraphics[width=\picScaleFr\columnwidth]{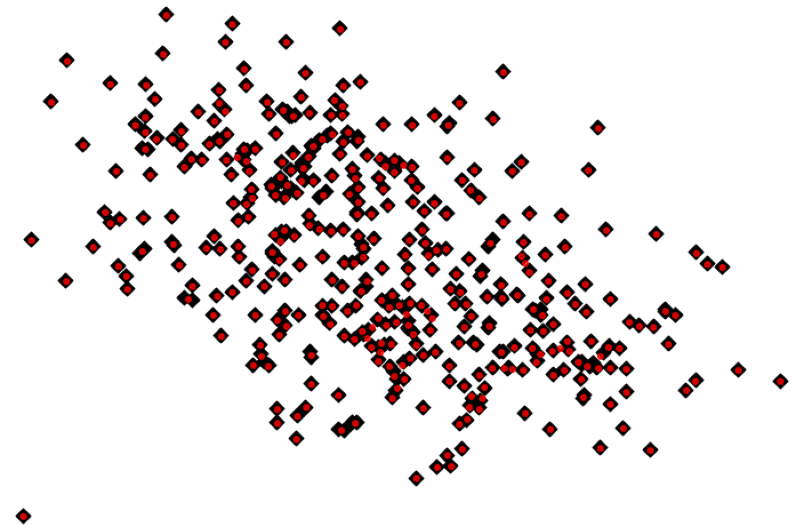}&
		\includegraphics[width=\picScaleFr\columnwidth]{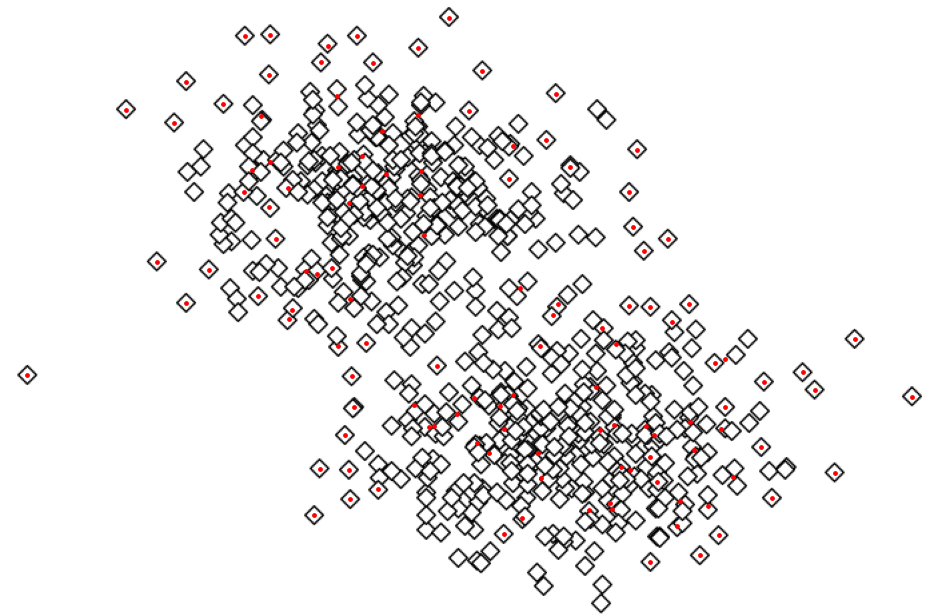}&
		\includegraphics[width=\picScaleFr\columnwidth]{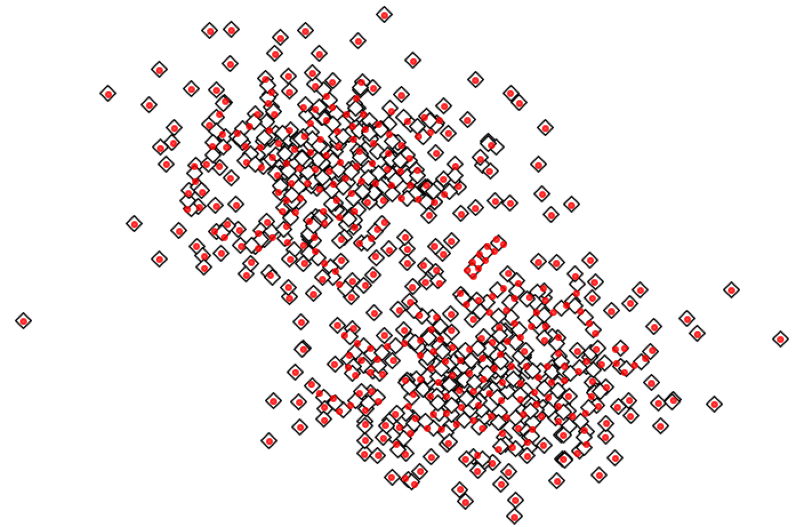}\\
		0.341&0.870&0.171&0.604\\
		\includegraphics[width=\picScaleFr\columnwidth]{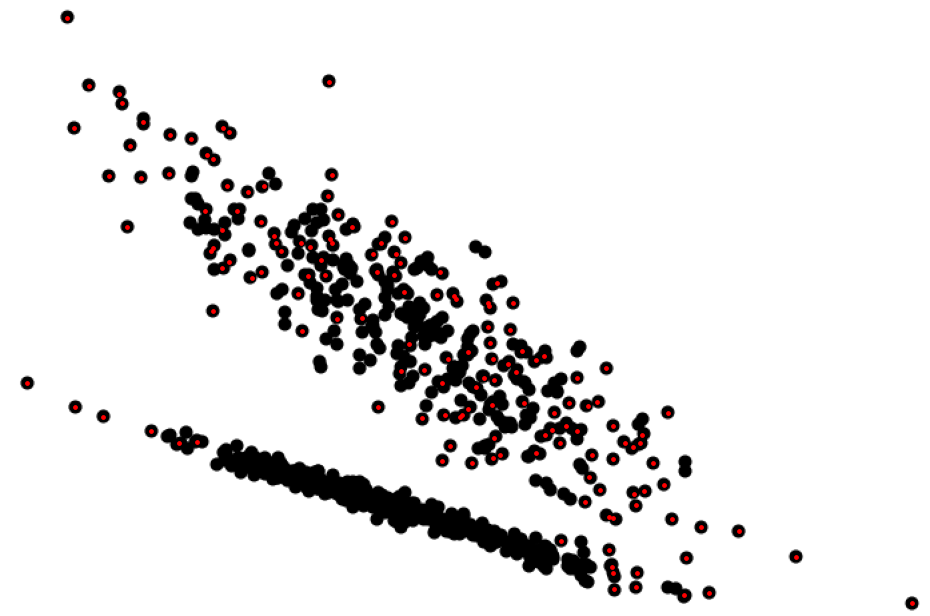}&
		\includegraphics[width=\picScaleFr\columnwidth]{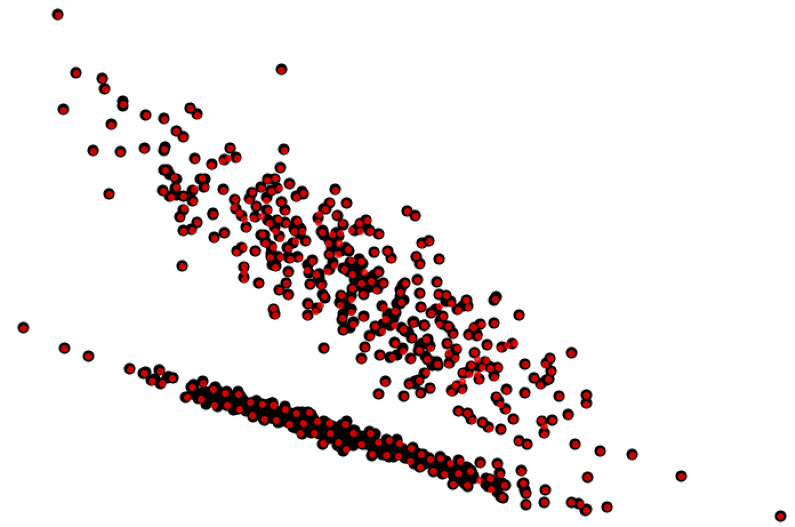}&
		\includegraphics[width=\picScaleFr\columnwidth]{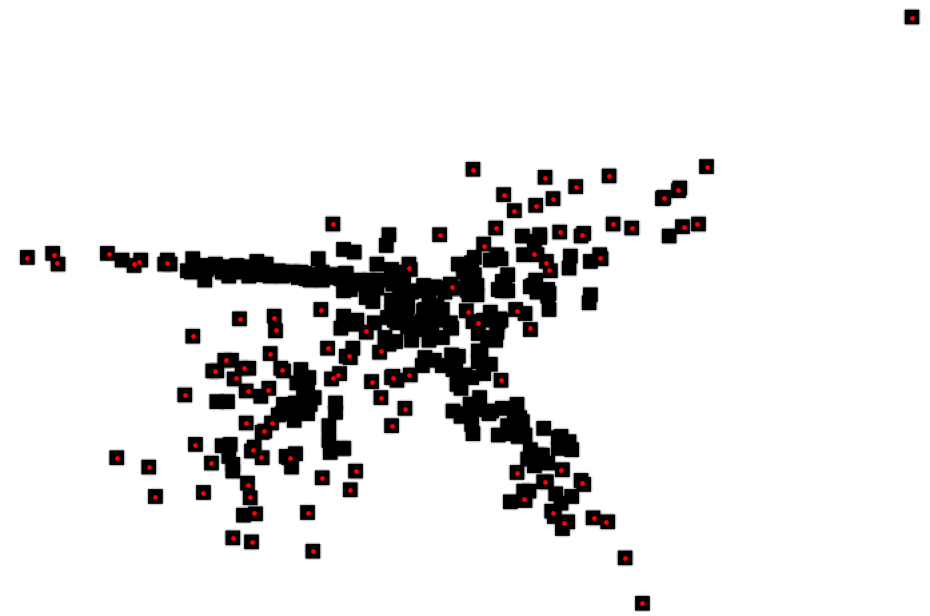}&
		\includegraphics[width=\picScaleFr\columnwidth]{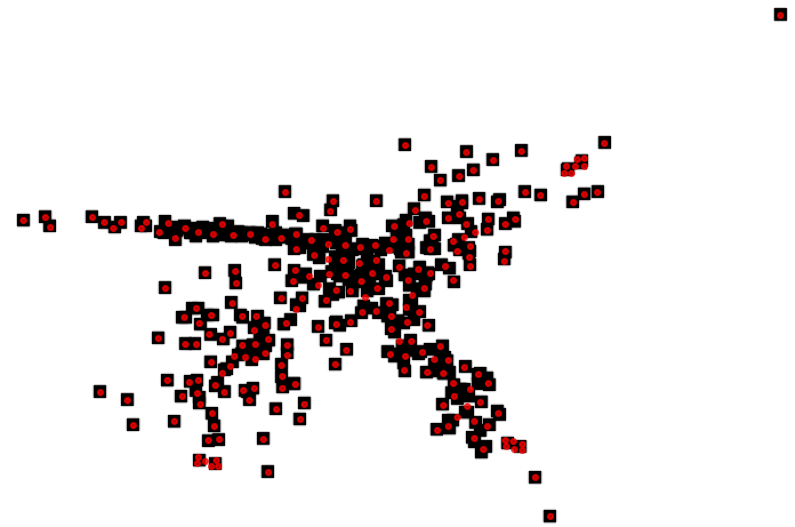}\\
		0.211&0.509&0.245&0.561\\
		\includegraphics[width=\picScaleFr\columnwidth]{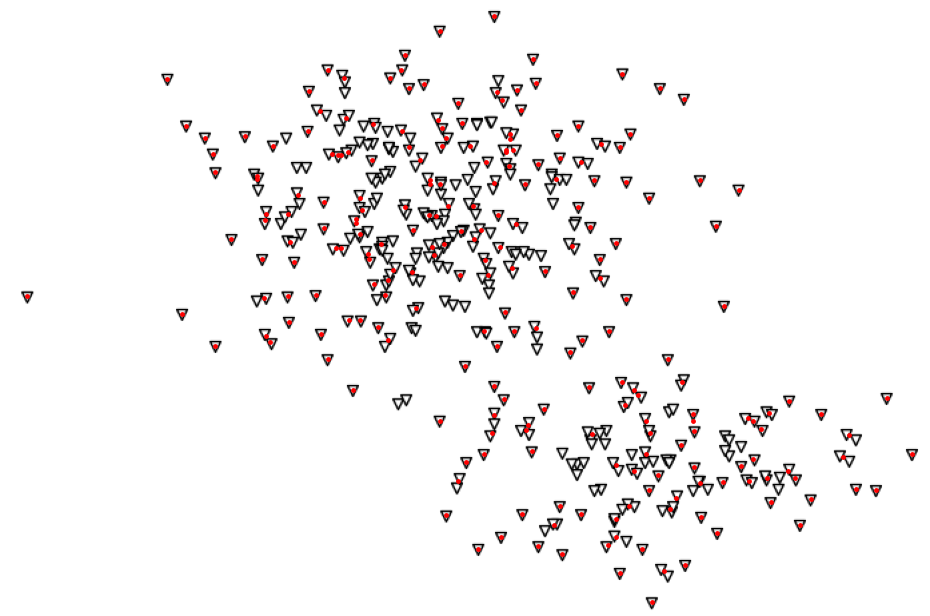}&
		\includegraphics[width=\picScaleFr\columnwidth]{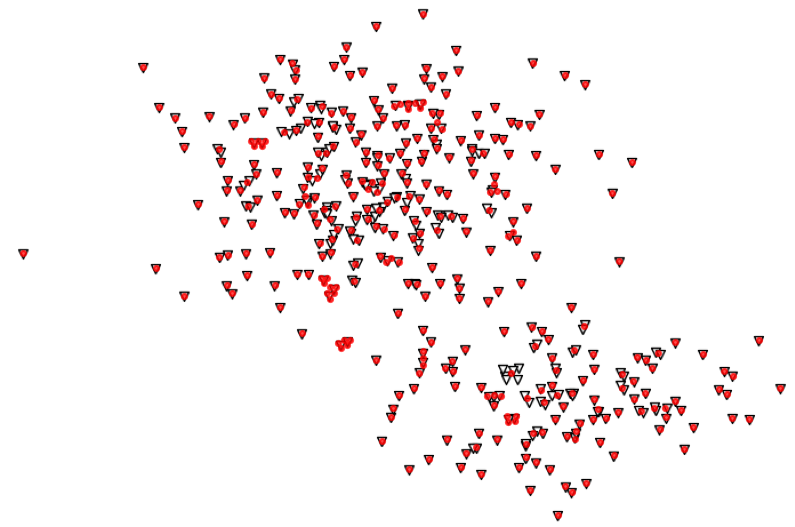}&
		\includegraphics[width=\picScaleFr\columnwidth]{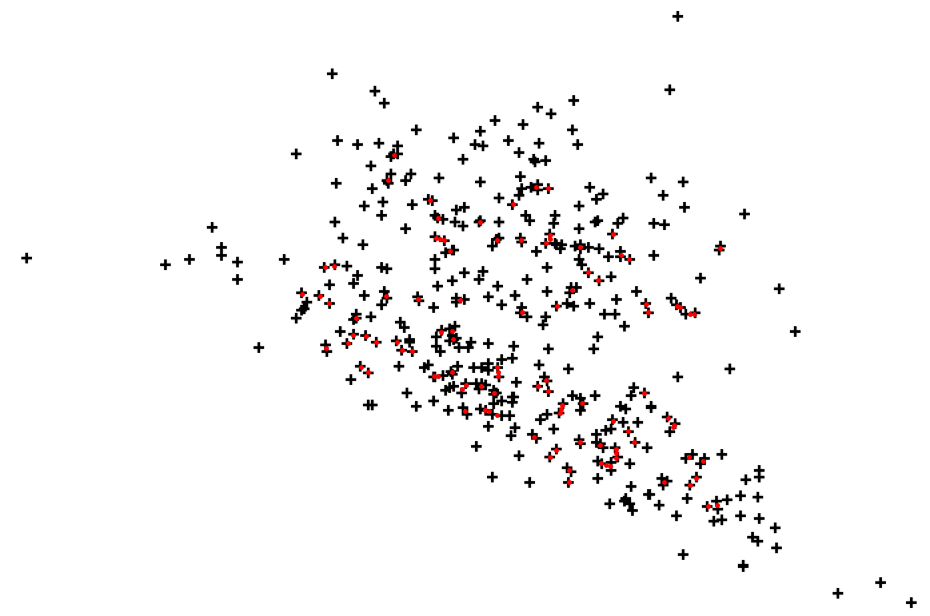}&
		\includegraphics[width=\picScaleFr\columnwidth]{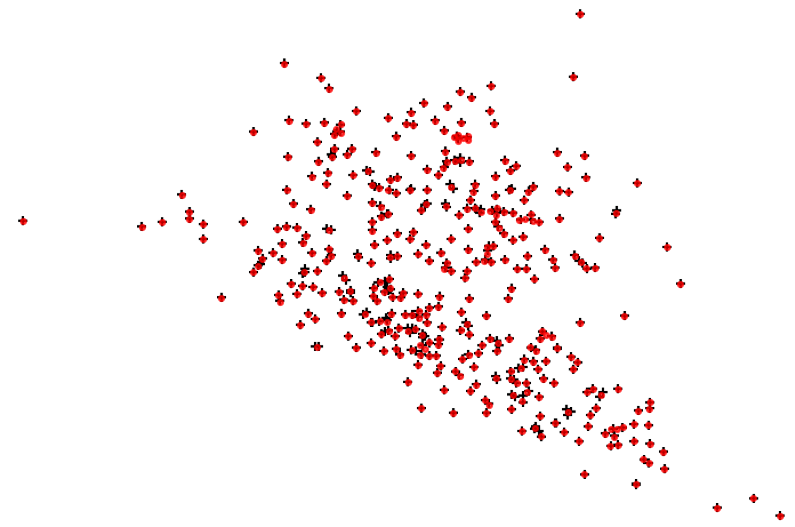}\\
		0.567&0.842&0.251&0.869\\
		ScatterScanner+RSMA&OsmLocator+RSMA&ScatterScanner+RSMA&OsmLocator+RSMA\\
	\end{tabular}
	\caption{Examples of experimental results on SML2023 dataset. Red points in images denote the located marks. The number under each image denotes the ACB score. \label{fig:expr_rsma}}
\end{figure}

Table \ref{tb_with_rsma} lists the experimental results with RSMA. It is shown that OsmLocator with RSMA obtain a good ACB score 77.9\%, which is 31.1\% higher than ScatterScanner with RSMA. It indicates that the proposed methods works well and RSMA is effective on SML2023 dataset. So, considering almost all of scatter images have isolated single marks, it is a feasible and good choice to use RSMA in practice. Fig. \ref{fig:expr_rsma} presents the localization results by OsmLocator and ScatterScanner. It is shown that the proposed method can locate most of all marks in scatter images of various types. In comparison with ScatterScanner, our method has prominent advantage to locate overlapping marks.

\begin{figure}[!h]
	\begin{center}
		\includegraphics[width=0.95\linewidth]{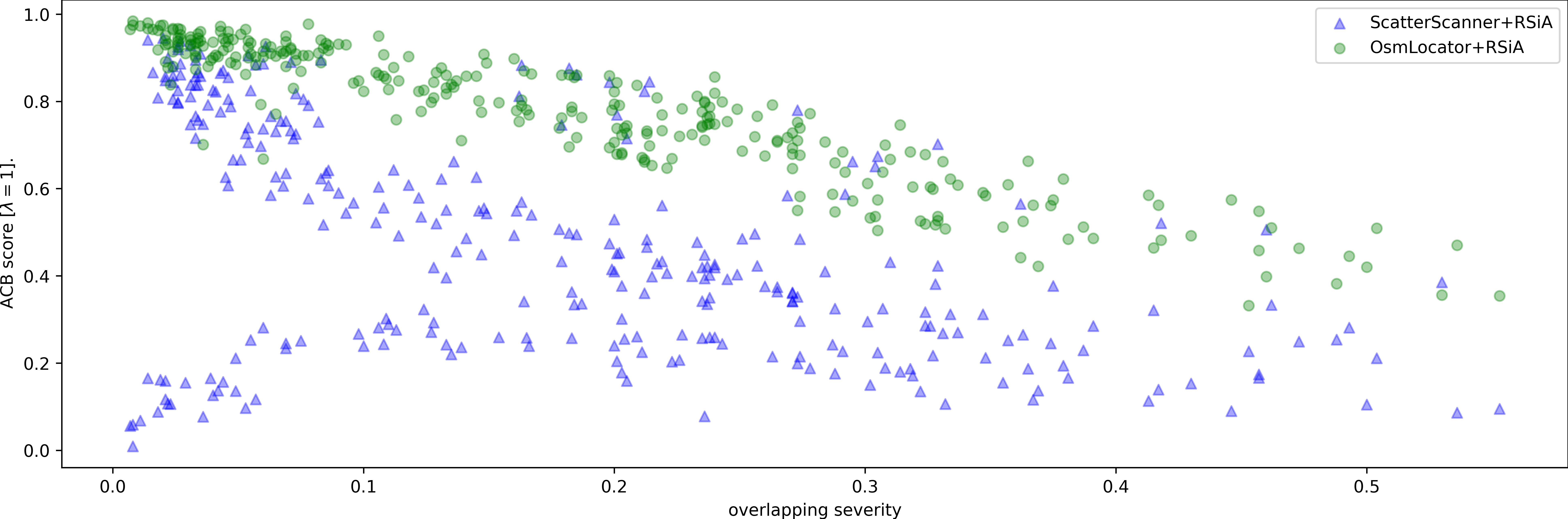}
	\end{center}
\caption{Scatter plot of the association between performance and overlapping severity.}
\label{fig:perf_os}
\end{figure}

A significant feature of SML2023 dataset \ref{subsec_dataset} is that we calculated the overlapping severity information for every scatter image, so we analysed the association between performance and overlapping severity. 
Fig. \ref{fig:perf_os} presents the analysed results, and it shows that OsmLocator obtain better performance than ScatterScanner on almost of all test cases. It is obvious that the performance of OsmLocator degrade gradually with the increase of overlapping severity.

\begin{figure}[!h]
	\begin{center}
		\includegraphics[width=0.95\linewidth]{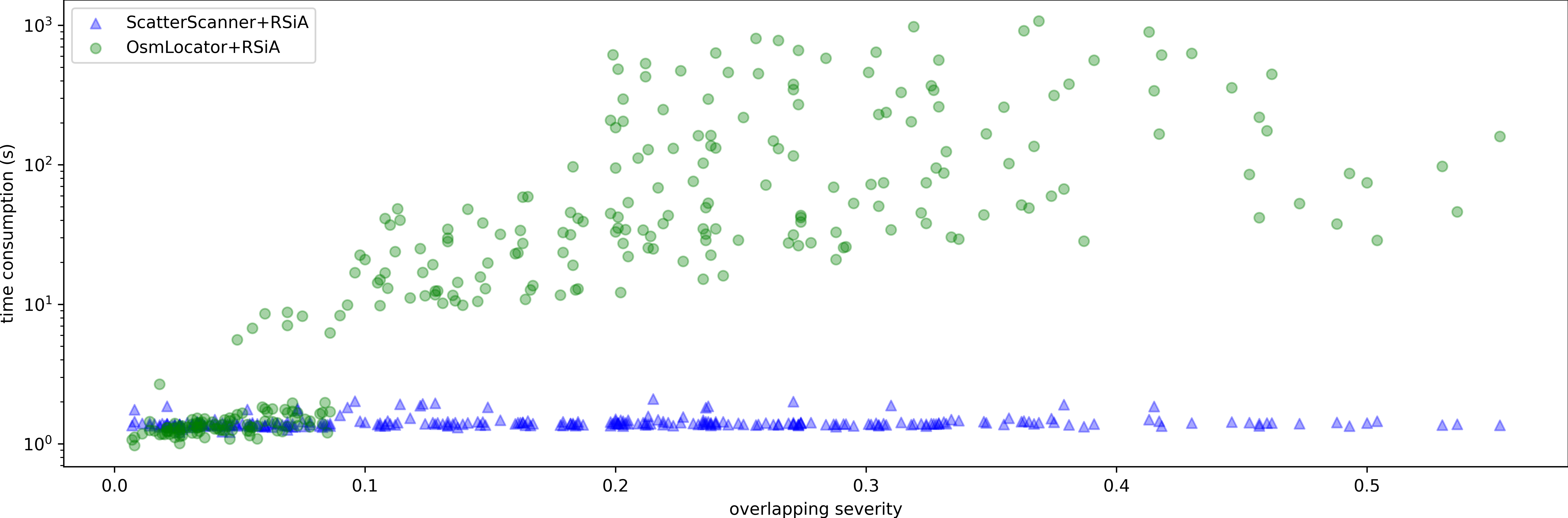}
	\end{center}
	\caption{Scatter plot of the association between time consumption and overlapping severity.}
	\label{fig:time_os}
\end{figure}

\begin{figure}[!h]
	\begin{center}
		\includegraphics[width=0.95\linewidth]{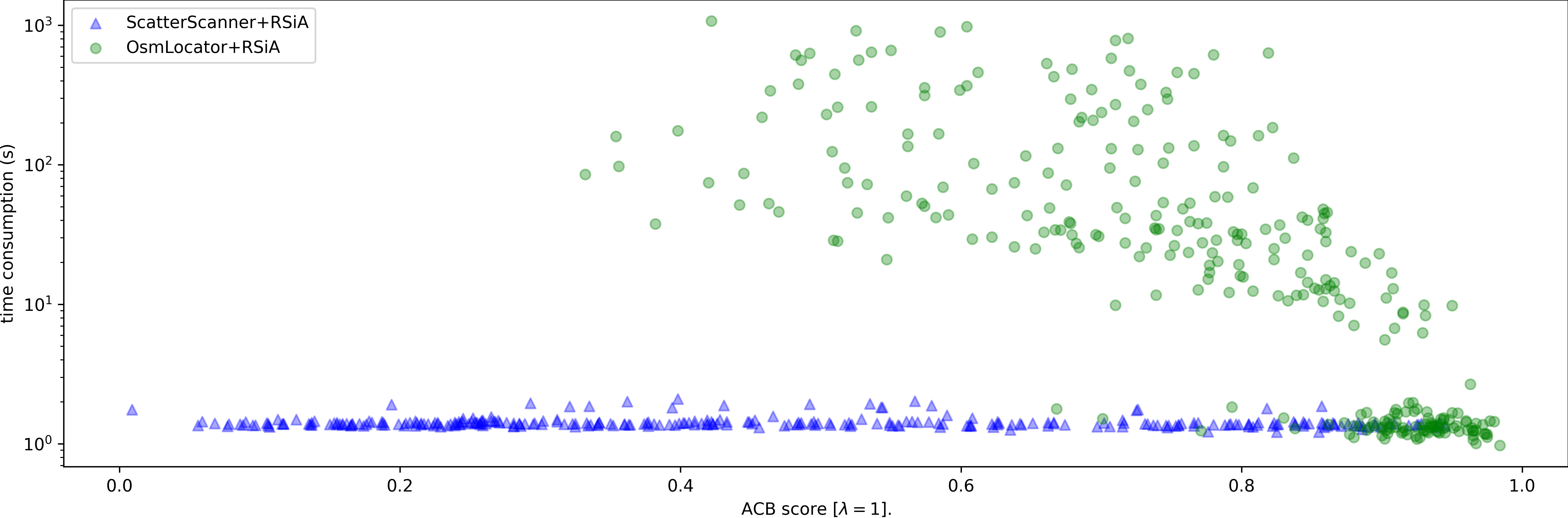}
	\end{center}
	\caption{Scatter plot of the association between performance and time consumption.}
	\label{fig:time_acb}
\end{figure}

Moreover, an inevitable problem we should analyse is the high computational cost of the proposed method, because the dependent K-Means clustering algorithm and SA algorithm both have high complexity. In order to observe the computational cost quantitatively, we recorded the time consumption of all test cases and analysed the relationship between time consumption and overlapping severity. Figure \ref{fig:time_os} presents the relationship between them. It is shown that the time consumption rapidly rises with the increase of overlapping severity. Further, we made program profile for some cases to analyse time consumption, and found that the Kmeans clustering function costed almost of all time, over 99\% for some cases with high overlapping severity. In addition, we also analyse the association between time consumption and performance. As shown in Figure \ref{fig:time_acb}, OsmLocator usually spends less time on the cases with better performance. By contrast, ScatterScanner spent nearly same time to handle every case, namely do not increase time consumption with the rise of overlapping severity. Therefore, the performance improvement we gained at the expense of more time consumption for heavy overlapping cases. However, thanks to the SA algorithm, almost of all cases can be solved in acceptable time consumption. We will discuss the efficiency of SA in Subsection \ref{sec:infl_sa}.

\subsection{Experiments with different control parameters} \label{sec:dcp}
As stated in Section \ref{sec:formulation}, the control parameters are basically stable and weakly affected by the input scatter images. It is important to validate this via experiments. Therefore, we made some experiments with different combination of control parameters, and their results are listed in Table \ref{tb_ctrl_params}. It is shown that different control parameters in reasonable ranges do not impact the final performance apparently. This indicates that the control parameters of the proposed method have considerable stability, and a parameter combination can be safely applied in different cases in different scenarios. In other words, the proposed method manifests the feasibility and effectiveness of self-adaption.

\begin{table}[!h]
	\begin{center}
		\setlength{\tabcolsep}{0.35em}
		{\caption{Experimental results with different control parameters}\label{tb_ctrl_params}}
		\begin{tabular}{lccccc}
			\hline
			Method & \makecell{Control parameters \\ $[\alpha,\beta,\gamma_s,\gamma_m]$} & ACB[$\lambda=1$] & ACB[$\lambda=5$] & ACB[$\lambda=10$] & Time consumption\\
			\hline			
			OsmLocator+RSMA & [$1.1,1.0,1.5,1.5$]   & 0.779$\pm$0.153 & 0.798$\pm$0.157 & 0.801$\pm$0.158 & 95.5$\pm$182.7 \\
			OsmLocator+RSMA & [$1.3,3.0,1.5,1.5$]   & 0.773$\pm$0.159 & 0.791$\pm$0.161 & 0.793$\pm$0.162 & 95.2$\pm$186.6 \\
			OsmLocator+RSMA & [$1.3,3.0,2.0,2.0$]   & 0.773$\pm$0.162 & 0.789$\pm$0.164 & 0.791$\pm$0.164 & 140.8$\pm$281.8 \\
			OsmLocator+RSMA & [$1.5,5.0,1.5,1.5$]   & 0.766$\pm$0.161 & 0.782$\pm$0.163 & 0.784$\pm$0.163 & 95.6$\pm$186.7 \\
			OsmLocator+RSMA & [$1.5,5.0,2.0,2.0$]   & 0.767$\pm$0.166 & 0.782$\pm$0.168 & 0.784$\pm$0.168 & 141.4$\pm$281.3 \\
			\hline
		\end{tabular}
	\end{center}
\end{table}

\subsection{Influence of control parameters}\label{sec:icp}
The control parameters ($\alpha, \beta, \mathfrak{F}$) in Eq.\ref{eq_singleOpt} are critical to obtain a good localization of marks in scatter images. How to find good values of control parameters is becoming an important problem in practice. Totally, The control parameters in Eq.\ref{eq_singleOpt} have specific and interpretable sense. The role of $\alpha$ is to make the single-objective has a global unique optimum, and it dominates the final results. Since the term $frac{n}{N_0}$ is adaptive for different scale inputs, a fixed value of $\alpha$ performs relatively stable for most cases. The role of $\beta$ is to make all marks in a binary region as equal as possible, and it only fine-tunes the final results for the cases that have a same mark size. It is also stale for most cases.

\begin{figure}[!h]
	\centering
	\setlength{\tabcolsep}{0.5em}
	\begin{tabular}{cccc}
		\includegraphics[width=0.15\textwidth]{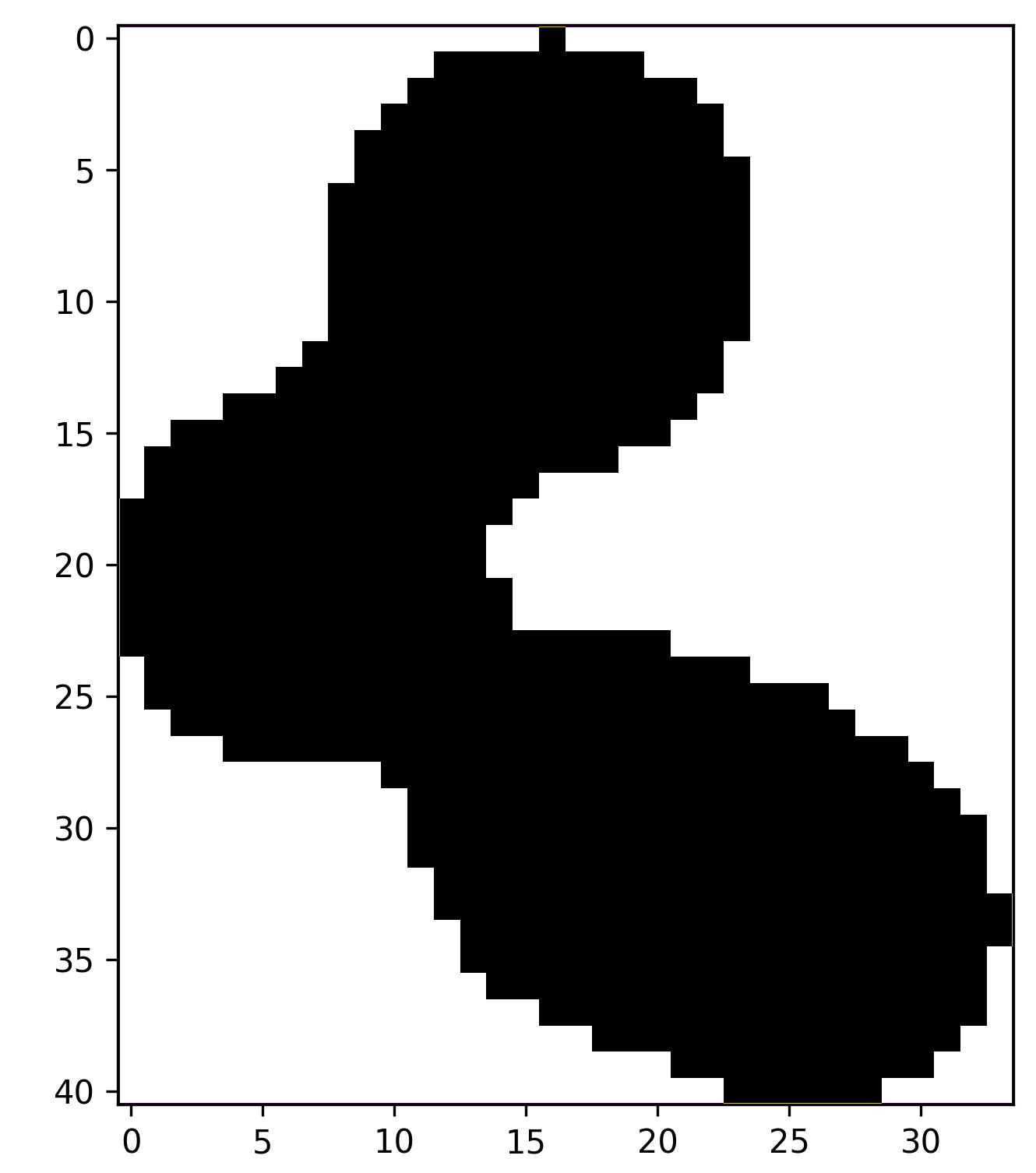} &
		\includegraphics[width=\picScaleEc\textwidth]{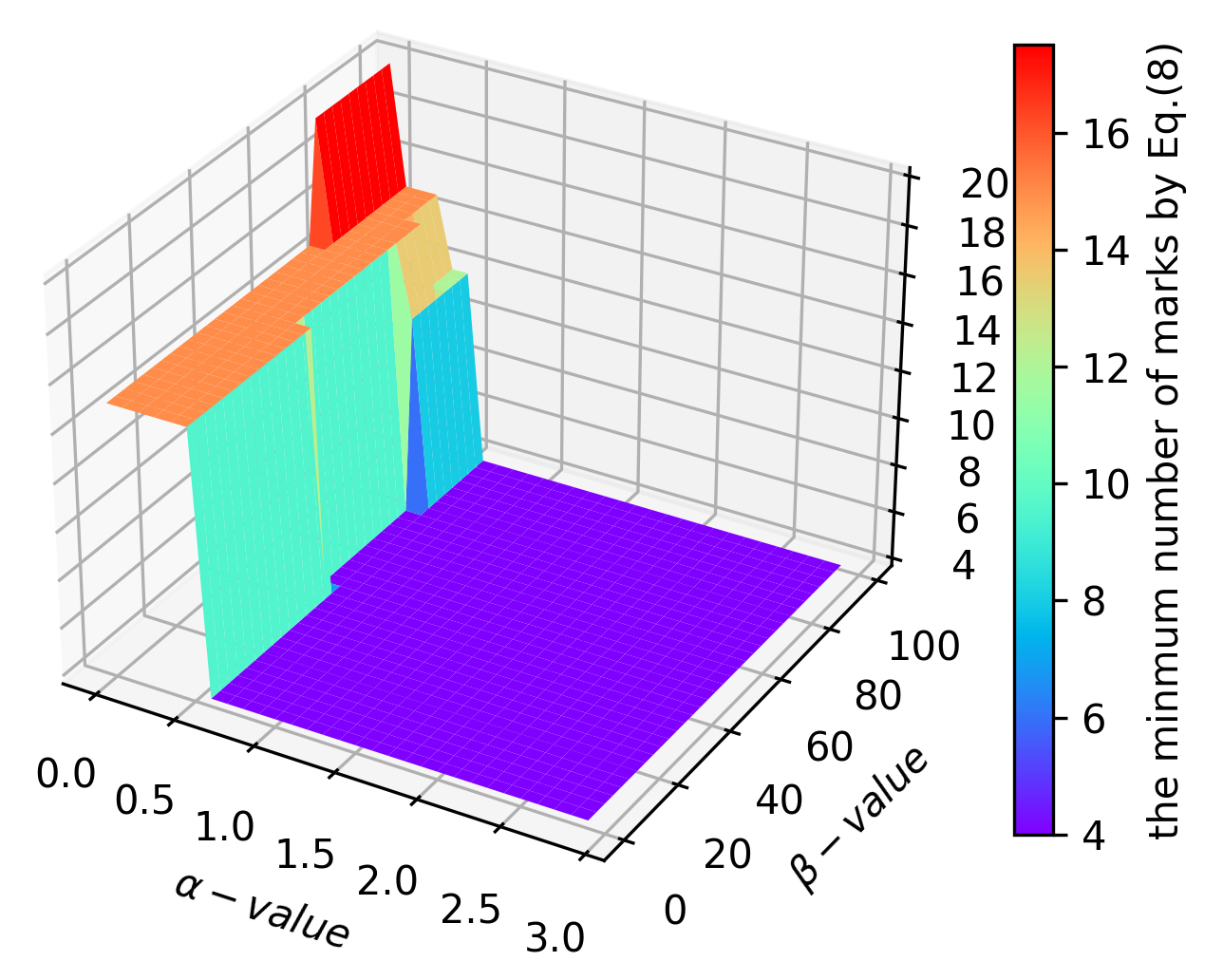} &
		\includegraphics[width=\picScaleEc\textwidth]{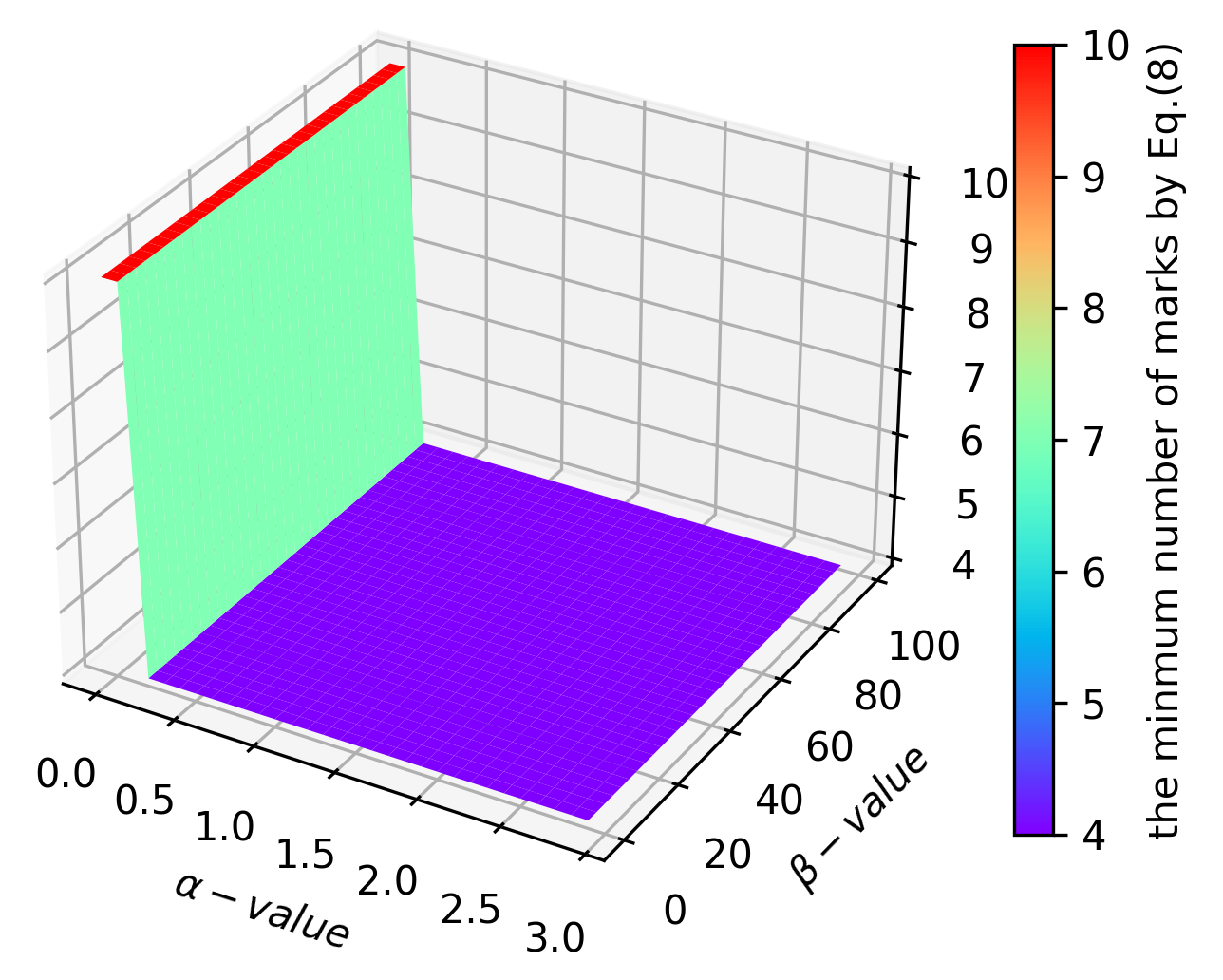} &
		\includegraphics[width=\picScaleEc\textwidth]{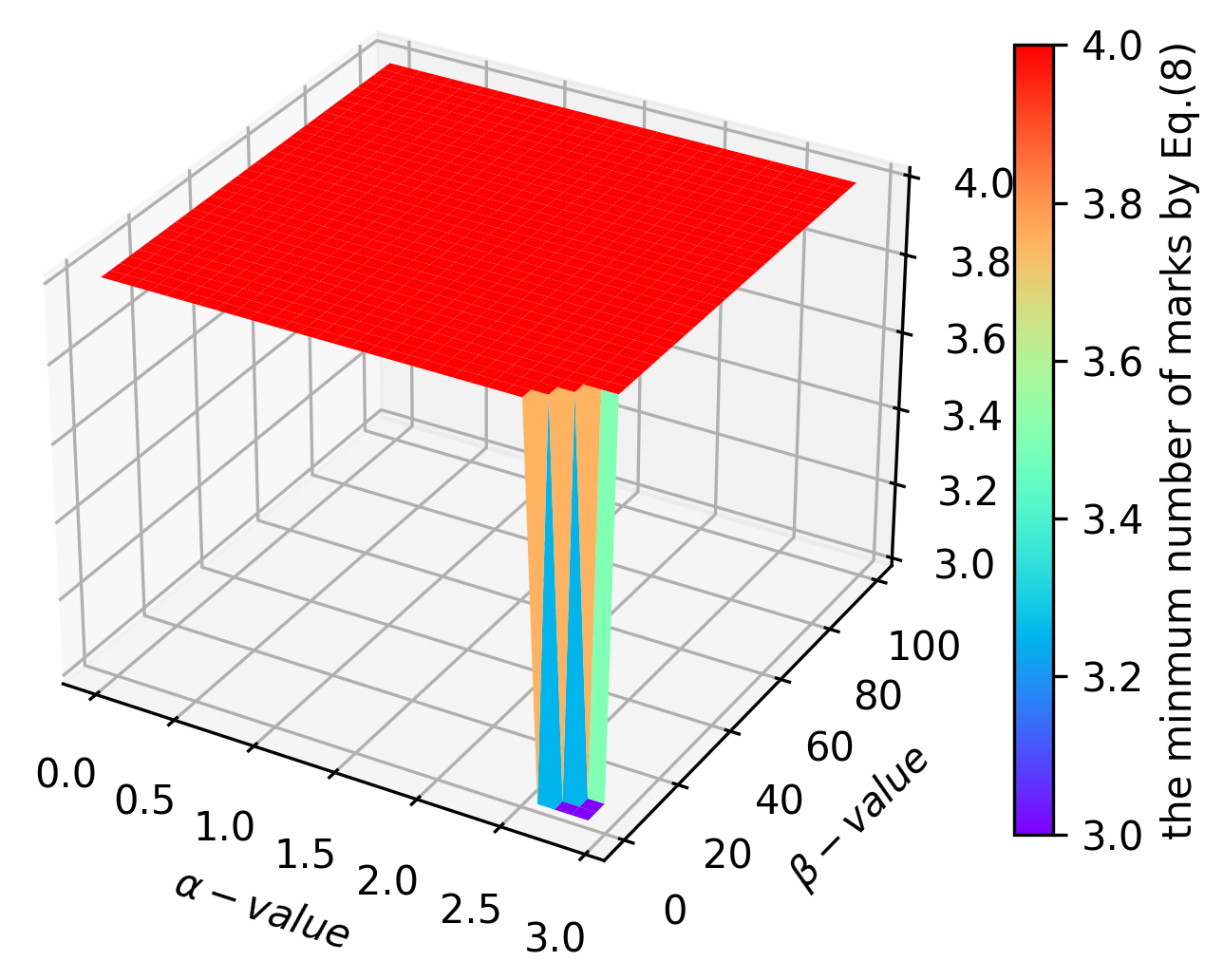} \\
		\includegraphics[width=0.15\textwidth]{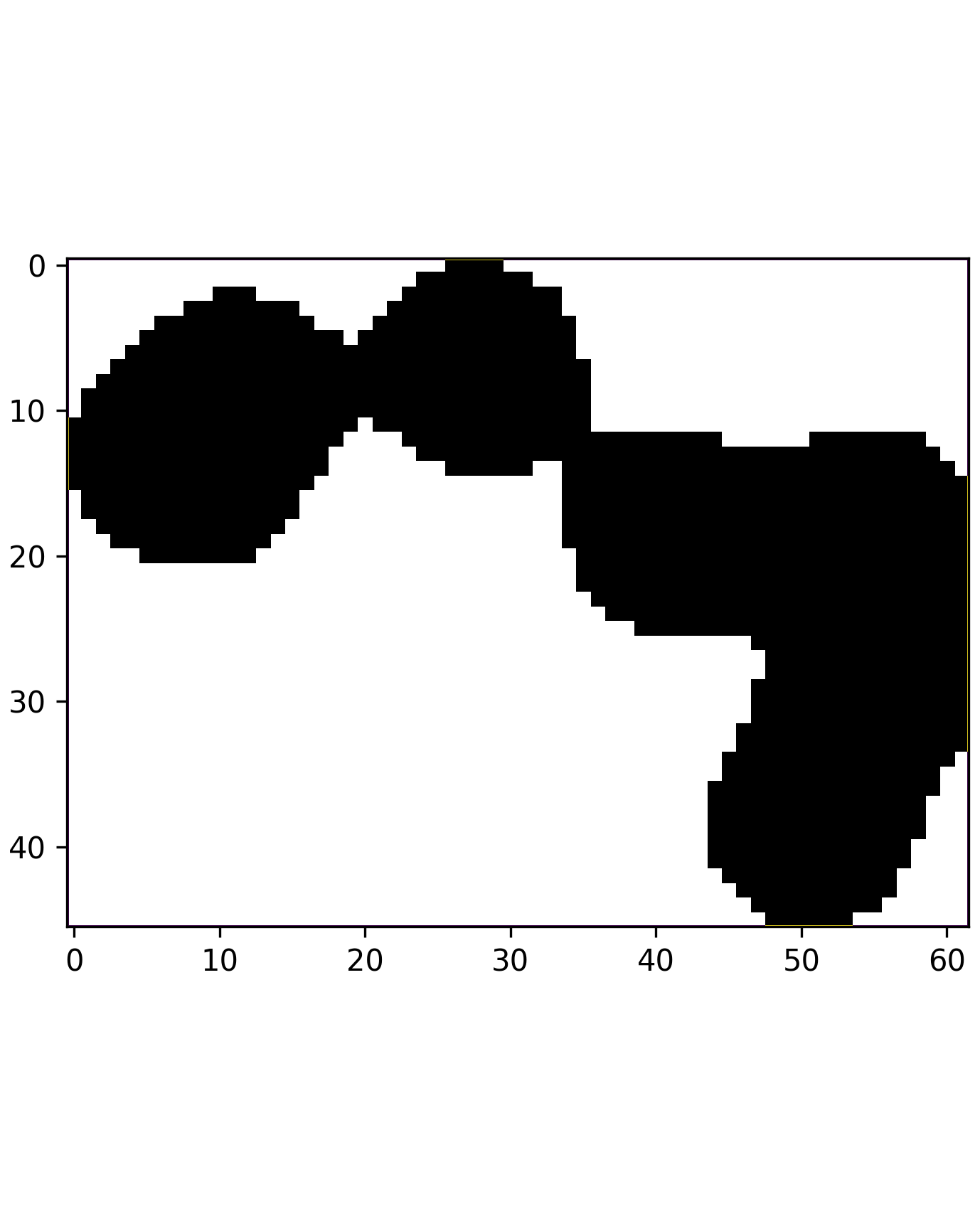}  &
		\includegraphics[width=\picScaleEc\textwidth]{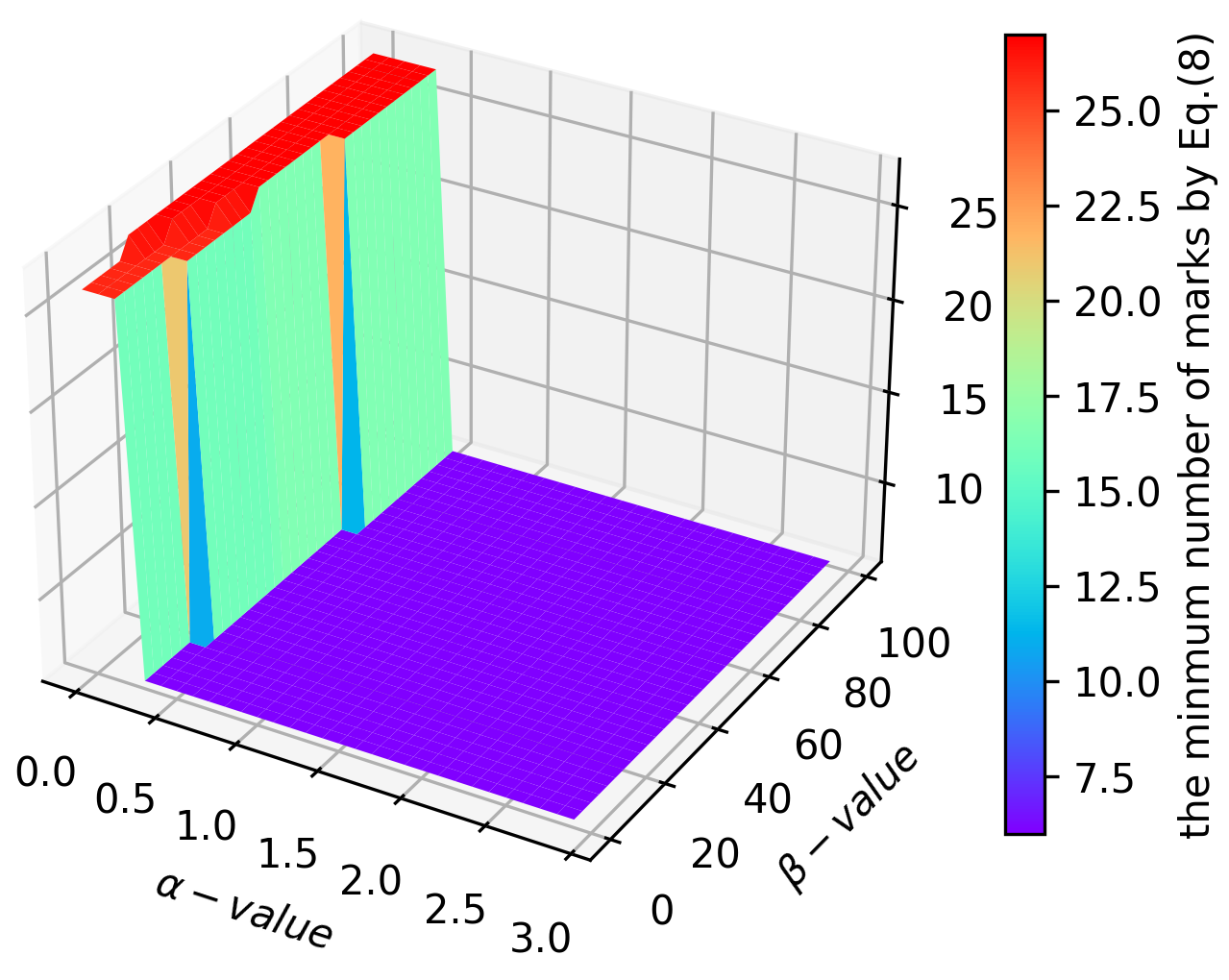} &
		\includegraphics[width=\picScaleEc\textwidth]{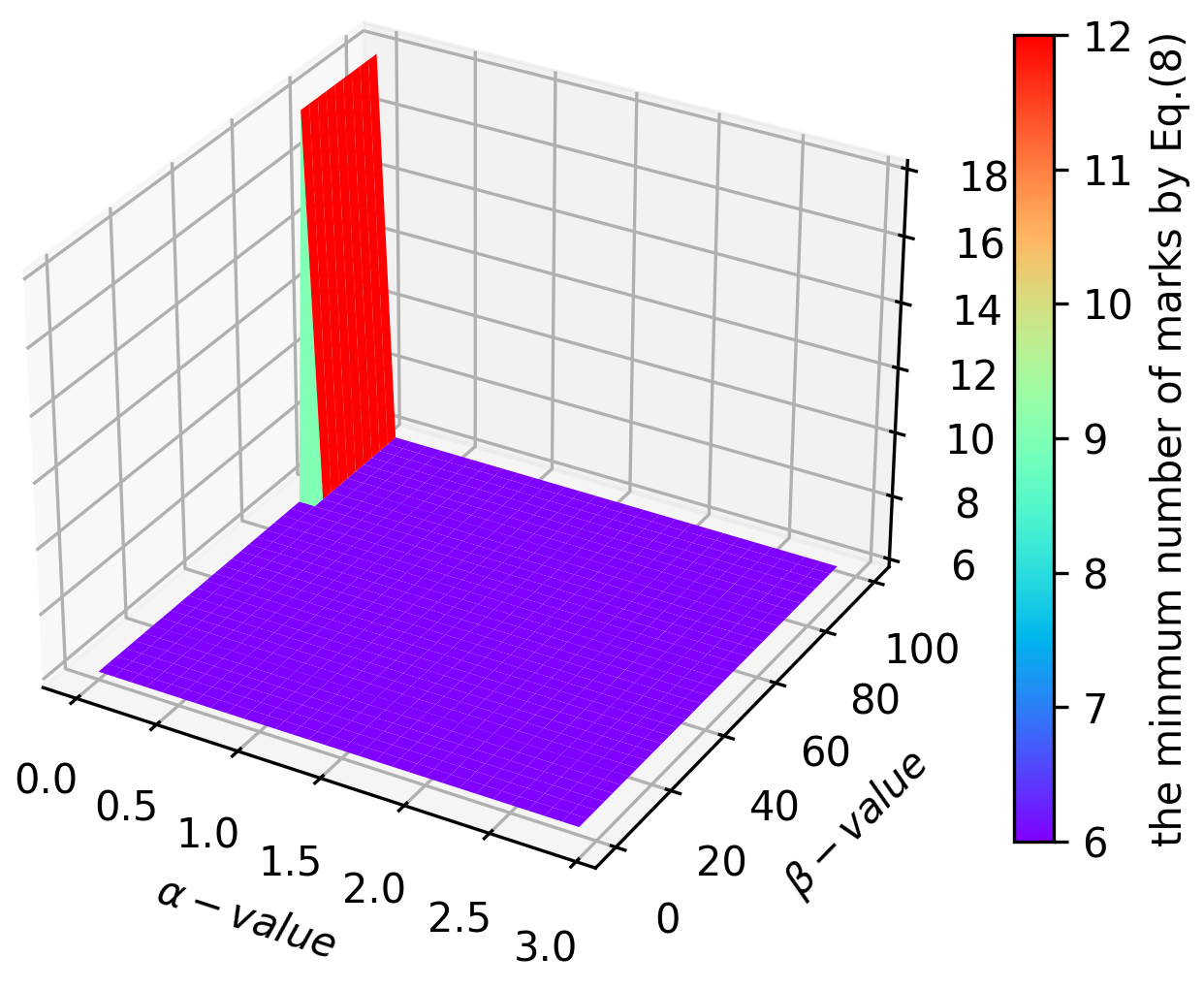} &
		\includegraphics[width=\picScaleEc\textwidth]{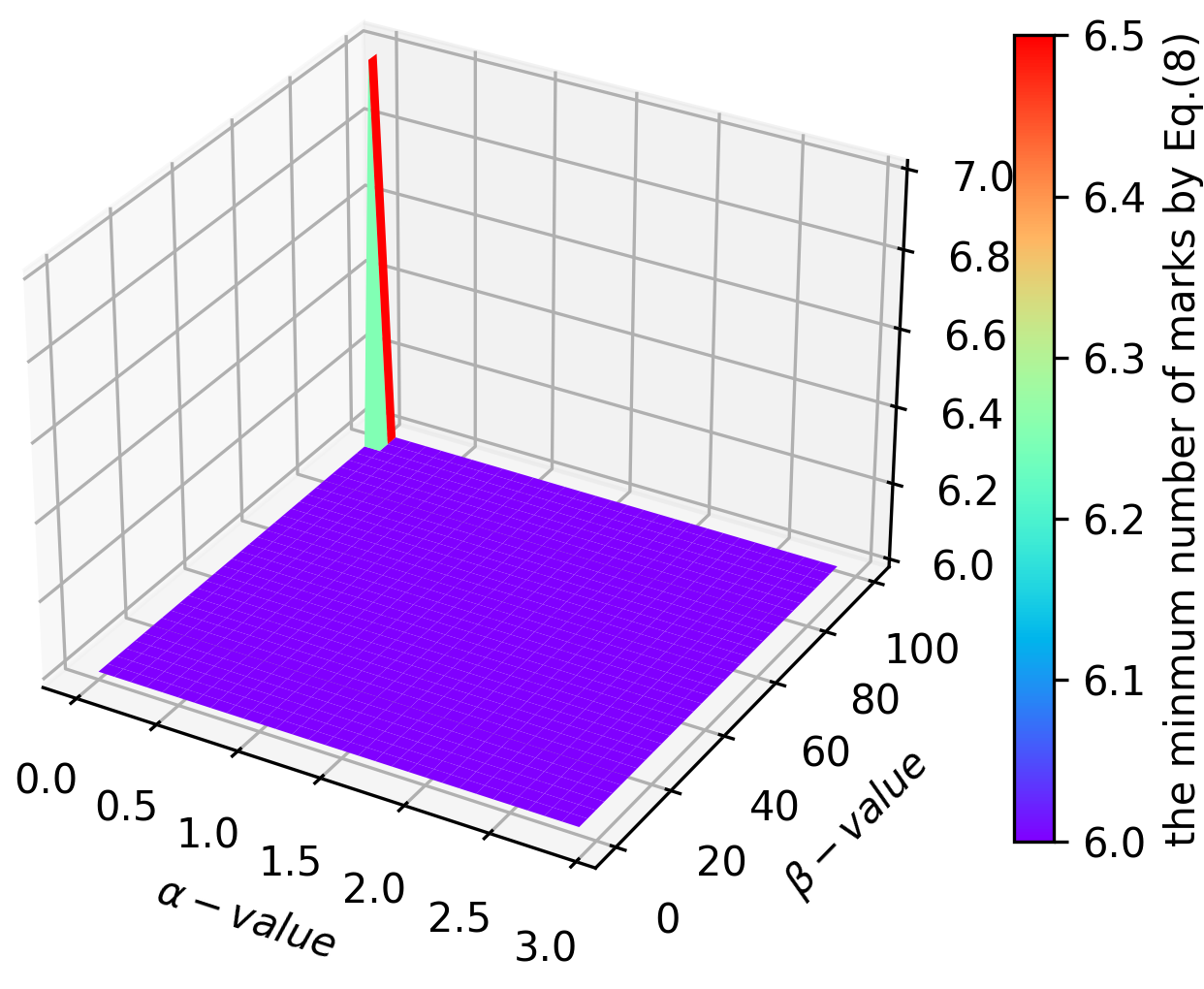} \\
		$\mathbb{C}$ & $\mathfrak{F}=30$ & $\mathfrak{F}=60$ & $\mathfrak{F}=90$ 
	\end{tabular}
	\caption{Two examples for showing the influence of control parameters $\alpha$ and $\beta$ on the estimated number of marks.}
	\label{fig:expr_infl_ab}
\end{figure}

Although experiments in subsection \ref{sec:dcp} have shown the stability of control parameters, it is still unknown the roughly-acceptable range of control parameters and how the control parameters influence results. Due to unacceptable time consumption of greedy algorithms, we only chose some quantitative experiments to investigate the influence of control parameters $\alpha$, $\beta$. 
In details, given two connected regions as shown in the first column of Figure \ref{fig:expr_infl_ab}, a set $S_\alpha$ was built from 0 to 3 with step 0.1, and a set $S_\beta$ was built from 0 to 100 with step 3; then the Cartesian product of $S_\alpha, S_\beta$ is $T=S_\alpha \times S_\beta=\{(\alpha_i, \beta_i)|\alpha_i \in S_\alpha, \beta_i \in S_\beta\}$; next, for each element in $T$, we used greedy algorithm to calculate the number of marks when $\mathcal{L}(\mathbb{C},n,m)$ reaches the minimum. Figure \ref{fig:expr_infl_ab} presents the experimental results of two examples. It is shown that the minimum of $\mathcal{L}(\mathbb{C},n,m)$ with most of all pairs of $T$ are right, so this indicates that our formulation of this problem is feasible and control parameters have a wide range to perform stably. Moreover, the space setting factor decides the size of searching space of solutions, and has weak influence on the setting of $\alpha$ and $\beta$ because same $\alpha$ and $\beta$ with different $\mathfrak{F}$ almost result in same performance.

\subsection{Effectiveness and Efficiency of simulated annealing} \label{sec:infl_sa}
\begin{figure}[!h]
	\centering
	\begin{subfigure}{0.247\columnwidth}
		\includegraphics[width=\textwidth]{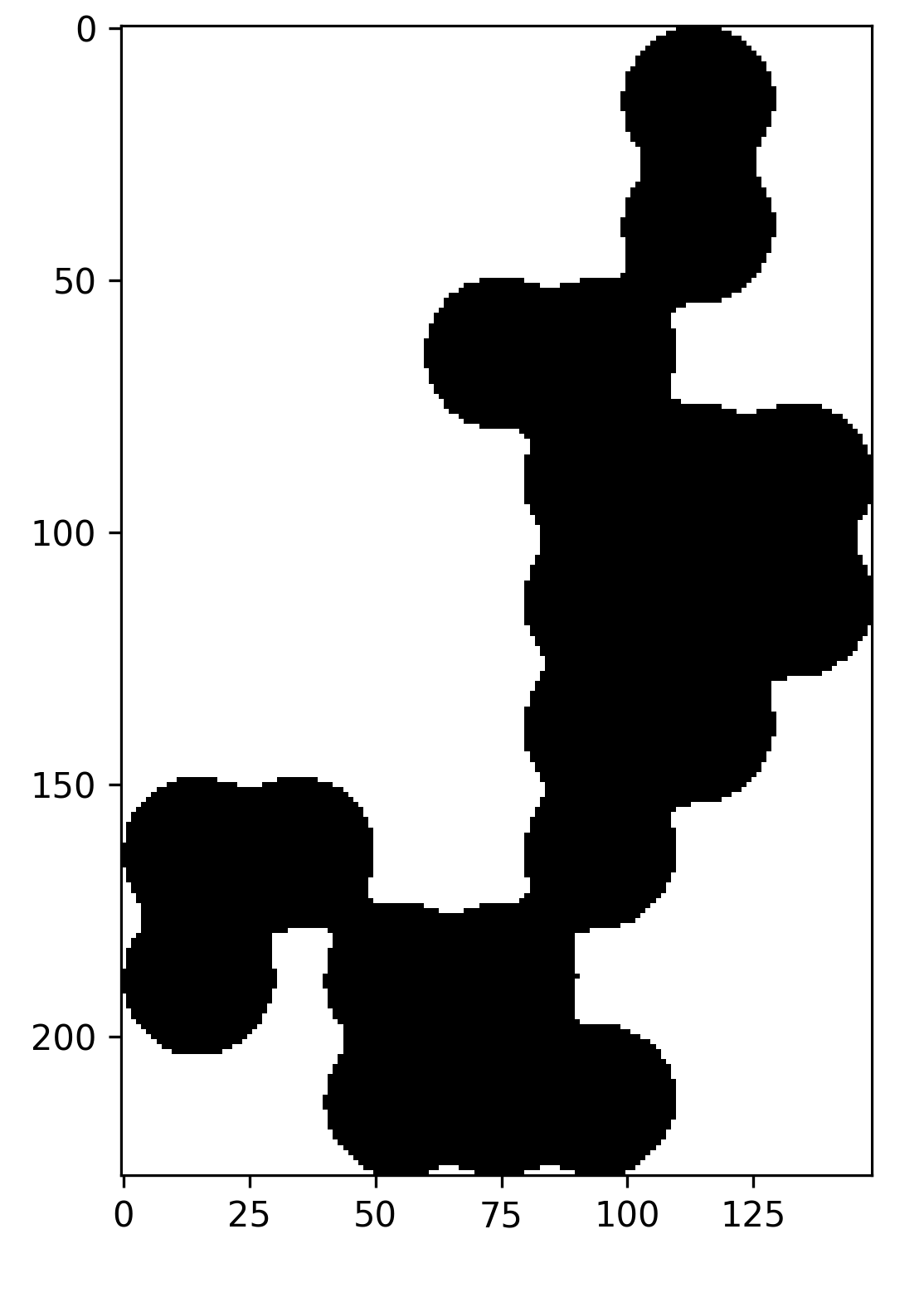}
		\caption{A connected region $\mathbb{C}$}
	\end{subfigure}
	\hfill
	\begin{subfigure}{0.33\columnwidth}
		\includegraphics[width=\textwidth]{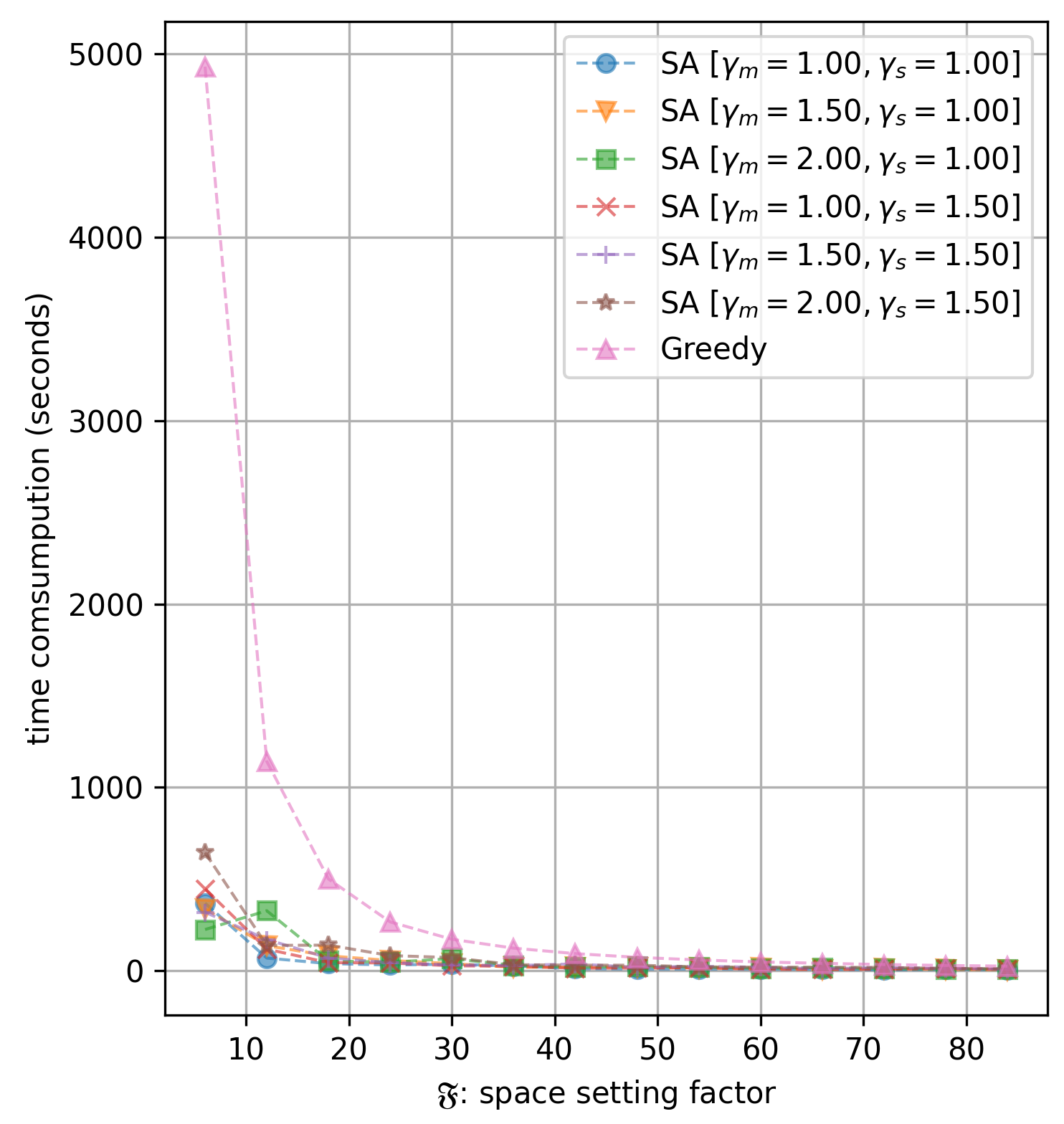}
		\caption{Time consumption.}
	\end{subfigure}
	\hfill
	\begin{subfigure}{0.33\columnwidth}
		\includegraphics[width=\textwidth]{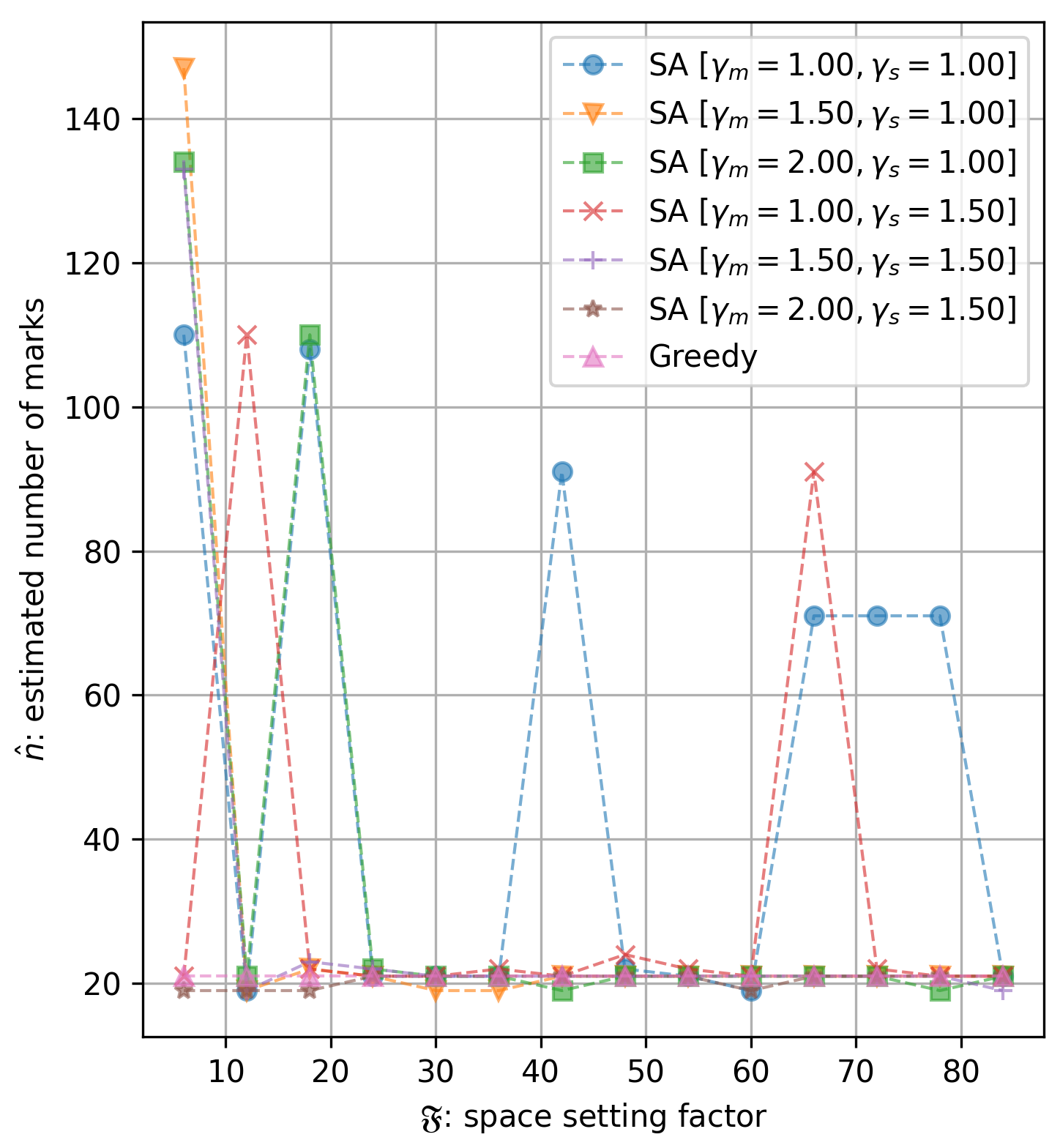}
		\caption{Estimated number of marks.}
	\end{subfigure}
	\caption{An case showing the performance of simulated annealing with difference control parameters. It is shown that simulated annealing is effective and efficient in appropriate settings.}
	\label{fig:expr_sa}
\end{figure}

In this work, we employ simulated annealing to obtain the optimal solution. In order to make clear whether the simulated annealing is effective and what is the efficiency of simulated annealing, we made some experiments in comparison with greedy algorithm on a connected region. In details, for fixed control parameters $\alpha=1.5$ and $\beta=5$, we took a list of space setting factors from 6 to 90 with step 6, and then recorded the time consumption and estimated number of marks for different pairs of parameters of simulated annealing, namely the coefficient of stop criteria $\gamma_s$ and coefficient of Markov steps $\gamma_m$. According to the experimental results in Figure \ref{fig:expr_sa}, we can come to three conclusions as follows. First, the simulated annealing is effective and efficient for optimization of the formulation we defined in section \ref{sec:formulation}. The simulated annealing can stably give us a acceptable solution in most cases, but run much faster than greedy algorithm, more than 95\% reduction of time consumption on average. Second, in contrast to the stop criteria defined in Eq. \ref{eqGammaS}, Markov steps defined in Eq. \ref{eqGammaM} is more sensitive to the final performance. For an example, the estimated number of marks goes wrong in a smaller search space $\mathfrak{F}=66$ when $\gamma_m$ equals 1.0 even with a greater $\gamma_s=1.5$. Third, our formulation of this problem in section \ref{sec:formulation} performs good in almost all of search spaces, because the results of greedy algorithm are same in all $\mathfrak{F}$ values.

\section{Discussion} \label{sec_discz}
\begin{figure}[!h]
	\centering
	\begin{subfigure}{0.247\columnwidth}
		\includegraphics[width=\textwidth]{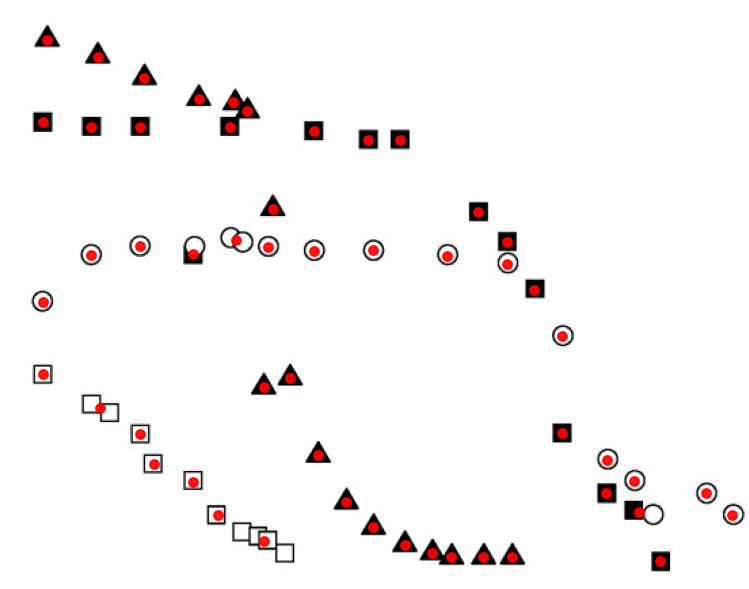}
	\end{subfigure}
	\hfill
	\begin{subfigure}{0.33\columnwidth}
		\includegraphics[width=\textwidth]{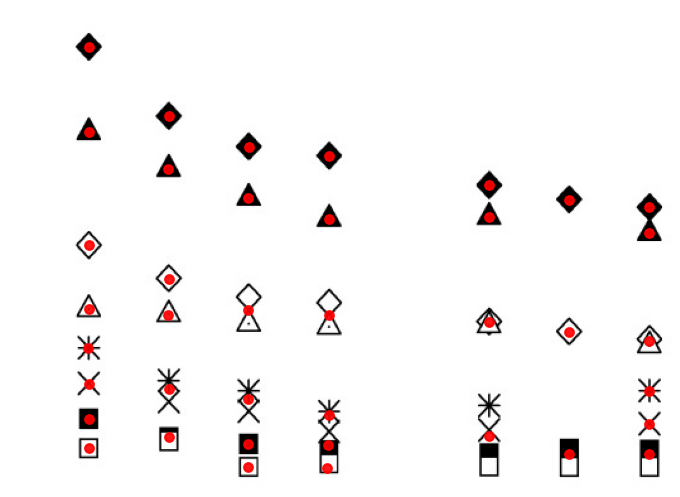}
	\end{subfigure}
	\hfill
	\begin{subfigure}{0.33\columnwidth}
		\includegraphics[width=\textwidth]{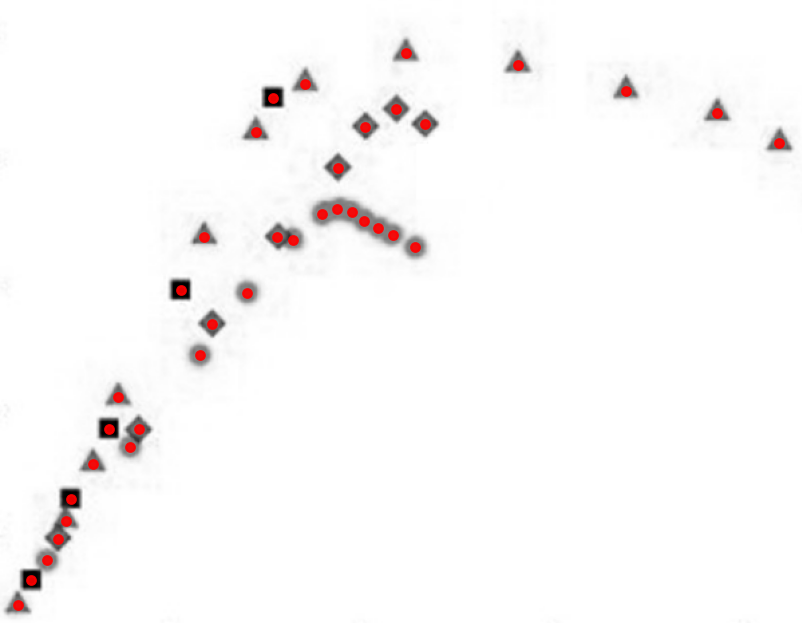}
	\end{subfigure}
	\caption{Three examples of overlapping marks localization with different marker shapes.}
	\label{fig:expr_diff_shapes}
\end{figure}
Our method still has some limitations. First, the proposed method does not have strong ability to locate the overlapping marks with different marker types. As shown in Figure \ref{fig:expr_diff_shapes}, our method failed to locate marks in some connected regions formed by different marker shape. A possible feasible way is to use different marker shapes to re-visualize every cluster, and we will improve it in this way in the future. As mentioned in Section \ref{subsec_dataset}, different markers usually be distinguished by different colours, so connected regions for different markers with same shape but different colours can be extracted by colour information and then use the proposed method to locate marks for each marker. This may be helpful to improve the performance further. Second, the employed simulated annealing is a Monte Carlo search method, so the time consumption is extremely high when too many marks overlap, it is possible to try some other optimization or other clustering method to reduce the time consumption in future works. Third, sometimes the randomness of K-means algorithm may affect the results, but if there are multiple runs (e.g. $n>=3$), it seldom affect the last results.
Although our method can recognise the type of markers, but we only focus on the task of mark localization in this work, and we will extend this work to support marker type recognition in future.

\section{Conclusion} \label{sec_conc}
Scatter images in various publications or documents are precious and valuable source of knowledge. Commonly-overlapping marks are difficult to locate through existing methods. 
This paper presents an unsupervised method to locate marks without any training dataset or reference. This method firstly uses K-means clustering algorithm to divide overlapping marks into clusters, and then utilizes the difference between re-visualization upon clusters and original visualization to formulate the task as an optimization problem, and employs simulated annealing method to solve the optimal solution and finally get each marker's position. We built a synthetic dataset SML2023 especially for evaluating overlapping marks localization, and tested the proposed method and comparable methods on SML2023. 
The experimental results show that the proposed method could accurately locate marks in many severely-overlapping scatter images, with about 30\% absolute increase on ACB metric in comparison with state-of-the-art methods.
This work can be applied in many applications, such as geographical coordinates extraction, information retrieval, business intelligence, data mining, publication digitalization.
Code sources and SML2023 dataset can be found in our github project \href{https://github.com/ccqym/OsmLocator}{https://github.com/ccqym/OsmLocator}.

\section*{Acknowledgements}
This research was funded by CAS Scholarship, and has been supported by the Flanders AI Research Programme grant No. 174B09119.

\bibliographystyle{elsarticle-num}
\bibliography{refs}

\end{document}